\pdfoutput=1

\documentclass[11pt]{article}

\usepackage{acl}

\usepackage{times}
\usepackage{latexsym}

\usepackage[T1]{fontenc}

\usepackage[utf8]{inputenc}

\usepackage{microtype}

\usepackage{inconsolata}

\usepackage[textwidth=1.7cm]{todonotes}
\iffalse
    \newcommand{\shikhar}[1]{}
    \newcommand{\henry}[1]{}
    \newcommand{\yc}[1]{}
\else
    \newcommand{\yc}[1]{{\bf \color{blue}{[YC: #1]}}}
    \newcommand{\shikhar}[1]{{\bf \color{red}{[ST: #1]}}}
    \newcommand{\henry}[1]{{\bf \color{orange}{[CH: #1]}}}
    
\fi

\usepackage{caption}
\usepackage{subcaption}
\usepackage[T1]{fontenc}
\usepackage[font=small,labelfont=bf,tableposition=top]{caption}

\DeclareCaptionLabelFormat{andtable}{#1~#2  \&  \tablename~\thetable}
\usepackage{amsmath,amsfonts,bm}
\usepackage{algorithm}
\usepackage{algpseudocode}
\usepackage{xspace}

\def\eqref#1{equation~\ref{#1}}

\def\1{\bm{1}}

\def\rvx{{\mathbf{x}}}

\DeclareMathAlphabet{\mathsfit}{\encodingdefault}{\sfdefault}{m}{sl}
\SetMathAlphabet{\mathsfit}{bold}{\encodingdefault}{\sfdefault}{bx}{n}

\usepackage{hyperref}
\usepackage{booktabs}
\usepackage{graphicx}
\usepackage{multirow}
\usepackage{caption}
\usepackage{amsmath}
\usepackage{amsthm}
\usepackage[english]{babel}
\usepackage[skins]{tcolorbox}
\usepackage{tikz}
\usepackage{color,soul}
\usepackage{enumitem}
\usepackage{float}

\usepackage{appendix}

\title{DynaMo: Accelerating Language Model Inference with Dynamic Multi-Token Sampling}

\def\@fnsymbol#1{\ensuremath{\ifcase#1\or \dagger\or \ddagger\or
   \mathsection\or \mathparagraph\or \|\or **\or \dagger\dagger
   \or \ddagger\ddagger \else\@ctrerr\fi}}

\author{Shikhar Tuli\textsuperscript{\rm 1,2}\thanks{Work done as an intern at Samsung Research America.}, 
    Chi-Heng Lin\textsuperscript{\rm 2},
    Yen-Chang Hsu\textsuperscript{\rm 2},
    Niraj K. Jha\textsuperscript{\rm 1},
    Yilin Shen\textsuperscript{\rm 2},
    Hongxia Jin\textsuperscript{\rm 2} \\
    \textsuperscript{\rm 1}Department of Electrical and Computer Engineering, Princeton University\\
    \textsuperscript{\rm 2}Samsung Research America\\
    \texttt{\{shikhar.tuli,chiheng.lin,yenchang.hsu,yilin.shen,hongxia.jin\}@samsung.com,}\\
    \texttt{jha@princeton.edu}
}

\begin{document}

\maketitle

\begin{abstract}
Traditional language models operate autoregressively, i.e., they predict one token at a time. Rapid explosion in model 
sizes has resulted in high inference times. In this work, we propose DynaMo, a suite of multi-token prediction language 
models that reduce net inference times. Our models \emph{dynamically} predict 
multiple tokens based on their confidence in the predicted joint probability distribution. We propose a lightweight
technique to train these models, leveraging the weights of traditional autoregressive counterparts. Moreover, we propose novel 
ways to enhance the estimated joint probability to improve text generation quality, namely co-occurrence weighted masking 
and adaptive thresholding. We also propose systematic qualitative and quantitative methods to rigorously test the quality of 
generated text for non-autoregressive generation. One of the models in our suite, DynaMo-7.3B-T3, achieves same-quality 
generated text as the baseline (Pythia-6.9B) while achieving 2.57$\times$ speed-up with only 5.87\% and 2.67\% parameter and training time 
overheads, respectively.
\end{abstract}

\section{Introduction}

Recent research has demonstrated the tremendous promise of large language models (LLMs) as competent artificial
intelligence (AI) assistants~\citep{touvron2023llama2}. This has led to their rapid and widespread adoption as 
chatbots in diverse applications, e.g., healthcare, e-commerce, education, etc. However, the high computational 
requirements of LLM training and inference and the use of massive closed-source corpora have restricted their 
development to a few laboratories. The increasing number of open-source LLMs, including Pythia~\citep{biderman2023pythia} and LLaMA-2~\citep{touvron2023llama2}, 
democratizes research in natural language processing (NLP). For instance, Vicuna-13B~\citep{vicuna2023}, an 
instruction-finetuned LLaMA model~\citep{touvron2023llama}, has gained significant interest among researchers due to 
its exceptional instruction-following capabilities for its relatively compact size. Nevertheless, access and study of 
LLMs remain limited due to challenges involved in their efficient evaluation on resource-constrained devices.

\begin{figure*}
    \centering
    \includegraphics[width=0.8\linewidth]{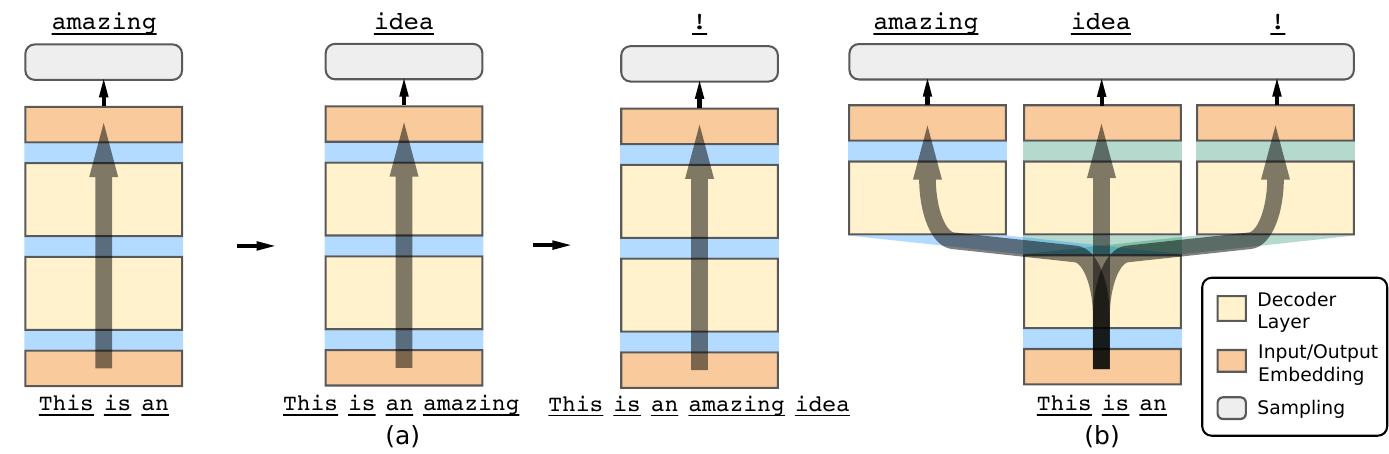}
    \caption{Multi-token prediction in DynaMo. (a) Traditional autoregressive prediction requires 
three forward passes. (b) Non-autoregressive multi-token prediction requires only one forward pass.}
    \label{fig:multi_token_intro}
\end{figure*}

\subsection{Challenges and Motivation}

LLM training and inference are typically limited to large GPU clusters in data centers, causing high latencies and privacy concerns for end-users. Edge computing offers a promising solution by processing data closer to the source, reducing latency and costs while enhancing data security and privacy. However, efficient deployment of conversational AI agents on resource-constrained edge platforms remains challenging, as even compact language models result in significant latencies~\citep{hanruiwang2020hat, tuli2023edgetran}. Increasing model sizes exacerbates this issue~\citep{kaplan2020scaling}, highlighting the need for significant inference/text-generation speed-ups and a range of models tailored to diverse platforms with varying resource constraints.

Existing models, trained with the causal language modeling (CLM) objective, predict one token at a 
time~\citep{radford2019gpt2, brown2020gpt3}. We conceptualize such models as $V$-way ($V$ is the vocabulary size) 
classifiers or unigram predictors. 
Mathematically, given the context, i.e., the set of past tokens $\rvx_{1:t} := \rvx_1, \rvx_2, \ldots, \rvx_t$, 
traditional LLMs model the probability distribution $p(\rvx_{t+1} | \rvx_{1:t}) = f_\theta(\rvx_{1:t})$, 
where $f_\theta$ is the LLM parameterized by $\theta$. In this context, traditional models generate sequences of text 
autoregressively. In other words, we sample $\rvx_{t+1}$ from $f_\theta(\rvx_{1:t})$ and then concatenate it with the 
input sequence to produce $\rvx_{1:t+1} := \rvx_1, \rvx_2, \ldots, \rvx_t, \rvx_{t+1}$. Then, we sample $\rvx_{t+2}$ 
from the predicted distribution $f_\theta(x_{1:t+1})$. Fig.~\ref{fig:multi_token_intro}(a) shows a schematic of this 
process with existing autoregressive LLMs.

Research in psycholinguistics shows that humans do not necessarily think of words one at a time when articulating 
thought~\citep{sridhar2012cognition}; instead they employ a parallel network of cognitive and linguistic processes. 
In line with this, we propose predicting multiple tokens simultaneously to accelerate inference. By estimating 
$p(\rvx_{t+1:t+3} | \rvx_{1:t}) = f_\theta$ (now, a $V^3$-way classifier), we aim to achieve reliable multi-token 
prediction, potentially resulting in a 3$\times$ inference speed-up (assuming no latency overhead). However, simultaneous 
prediction of three tokens may compromise generation quality (we provide sample generations in Appendix~\ref{app:sample_generations}). Hence, there is 
a need to \emph{dynamically} back off to lower-order $n$-gram prediction when the model lacks confidence.

\subsection{Our Contributions}

In this work, we propose DynaMo: a suite of \underline{\smash{dyna}}mic multi-token prediction language 
\underline{mo}dels. We target inference speed-up by improving upon traditional LLMs in terms of model architecture, 
training methodology, and non-autoregressive decoding schemes. Further, we propose novel methods to evaluate 
multi-token prediction for the next generation of non-autoregressive models. More concretely, we summarize the 
contributions of this work next.
\begin{itemize}
    \item We augment the suite of Pythia~\citep{biderman2023pythia} models for multi-token prediction. We explore 
various architectures for multi-token prediction (label shifts, masking strategies, multi-token heads, etc.). 
Further, we devise efficient ways to train augmented versions of existing pre-trained LLMs for multi-token prediction.
    \item We propose novel ways to dynamically predict multiple tokens based on the current context and probabilities 
of predicted tokens. We model the joint probability distributions of predicted tokens and \emph{back off} to lower-order $n$-gram 
prediction when the joint probabilities are not above a given threshold ($\epsilon_b$). We propose \emph{co-occurrence weighted masking} and \emph{adaptive 
thresholding} to improve generated text quality.
    \item We perform rigorous experiments to evaluate the downstream performance of our proposed models. We show that training 
with our modified-CLM objective enhances the first token prediction quality as well. We evaluate the open-ended 
text generation quality of our models and its dependence on model size, desired speed-up, and multi-token prediction 
hyperparameters (e.g., $\epsilon_b$). In fact, this is the first non-greedy, non-batched-parallel-decoding work that proves to deliver same-quality
generation as the base model with systematic qualitative and quantitative tests.
\end{itemize}

The rest of the article is organized as follows. Section~\ref{sec:methodology} details the multi-token 
prediction methodology adopted in the DynaMo suite of models along with the proposed evaluation methods. Section~\ref{sec:results} presents the experimental results. Section~\ref{sec:discussion} 
discusses the implications of multi-token prediction and points out future work directions. Finally, 
Section~\ref{sec:conclusion} concludes the article.

\section{Background and Related Works}

Previous research explores various approaches to reduce token prediction latency in LLMs. It includes 
distillation~\citep{hinton2015distilling}, complexity reduction~\citep{wang2020linformer}, 
sparsification~\citep{jaszczur2021sparse}, quantization~\citep{shen2020q}, etc., to reduce model size or complexity, leveraging 
specialized hardware~\citep{tuli2023acceltran}. 
Other engineering solutions include Flash attention~\citep{dao2022flashattention} that reduces memory reads/writes. Recently, skeleton-of-thought decoding~\citep{ning2023skeleton} was proposed, wherein the LLM first generates the 
skeleton of the answer and then conducts batched decoding to complete the contents of each skeleton point in parallel.  

Speculative decoding~\citep{stern2018blockwise, chen2023speculative} is yet another approach that has gained recent prominence. It leverages a small draft model (which can be combined with the main model,~\citealt{cai2023medusa}) to anticipate the main model predictions and queries it for batch verification. The batch size depends on the targeted number of token positions in the future, for draft prediction, and the number of top-$k$ samples at each position. Despite attempts at improving inference efficiency~\citep{spector2023accelerating, liu2023online}, such methods incur high computational overhead due to high-batch operations and result in poor compute utilization (e.g., sparse tree attention used by~\citealt{cai2023medusa, spector2023accelerating}). For the greedy decoding scheme, such methods enable up to $n\times$ speed-up, however, at the cost of at least $n\times$ the compute. Instead, in this work, we propose a low-compute approach that directly maps the joint probability distribution and implements co-occurrence weighted masking and adaptive thresholding, obviating the need for batched verification. Further, Medusa~\citep{cai2023medusa} exploits simple feed-forward layers for draft prediction. This work explores various architectural modifications for draft prediction. Nevertheless, the abovementioned approaches are orthogonal to the proposed method and can be used in conjunction to further boost performance.


\section{Method}
\label{sec:methodology}

In this section, we discuss the implementation details of multi-token prediction in the DynaMo suite.

\subsection{Going Beyond One-token Prediction}
\label{sec:going_beyond_one_token_prediction}

We propose a modified-CLM objective for multi-token prediction,
\begin{equation}
\label{eq:clm_multi}
    \mathcal{L}_{\text{T}n} = - \frac{1}{N} \sum_{j=1}^N \sum_{t=1}^{L - n + 1} \log p(\rvx_{t+n}^j|\rvx_{1:t}^j)
\end{equation}
for the $n^{th}$-token head. Here,  $N$ is the number of sequences in the training set and the length of the
$j^{th}$ sequence is $L$. The first-token head predicts the labels shifted by one position. The second-token head predicts the labels shifted by two positions, and so on. Note that the above equation trains each token 
head to predict the tokens independently. We approximate the joint probability distribution using independent 
token predictions. We represent this mathematically as follows:
\begin{align}
\begin{split}
\label{eq:prob_joint_independent}
    p(\rvx_{t+1:t+n} | \rvx_{1:t}) &= \prod_{i=1}^n p(\rvx_{t+i} | \rvx_{1:t+i-1}) \\
    & \approx \prod_{i=1}^n p(\rvx_{t+i} | \rvx_{1:t}) = \prod_{i=1}^n f_\theta^i(\rvx_{1:t})
\end{split}
\end{align}
where $f_\theta^i(\rvx_{1:t})$ is the prediction by the $i$-th-token head in the DynaMo model. 

We use the Pythia~\citep{biderman2023pythia} suite of models as base models. All decoder layers up to the 
penultimate layer form the model ``stem'' (like the stem of a plant). The final decoder layer of the base model 
and the output embedding form the first-token-predicting head (or simply the first-token head). 
Fig.~\ref{fig:multi_token_intro} shows the data flow for the base model in blue. It assumes a base model with only 
two decoder layers. The first layer of the base model forms the stem for the DynaMo model, while the second layer 
is part of the first-token head. The other decoder layers (dataflows shown in green) are part of the second- and 
third-token heads. The output embeddings for these heads reuse the weights of that of the first head. Hence, the 
extra parameters for this three-token model are from only two extra decoder layers.

Thanks to the above weight transfer process, most weights (the model stem and the first-token head) in an 
initialized DynaMo model are already trained. Therefore, we train the DynaMo models on a much smaller dataset 
(5\% randomly sampled version of the Pile dataset,~\citealt{gao2020pile}) relative to that used to train the 
Pythia models. This limits the computational overhead of training our models. We provide further details on the training
and evaluation methods for our models in Appendix~\ref{app:training_and_evaluation_process}.

\subsection{Dynamic Text Generation}
\label{sec:dynamic_text_generation}

\begin{figure*}
    \centering
    \includegraphics[width=0.8\linewidth]{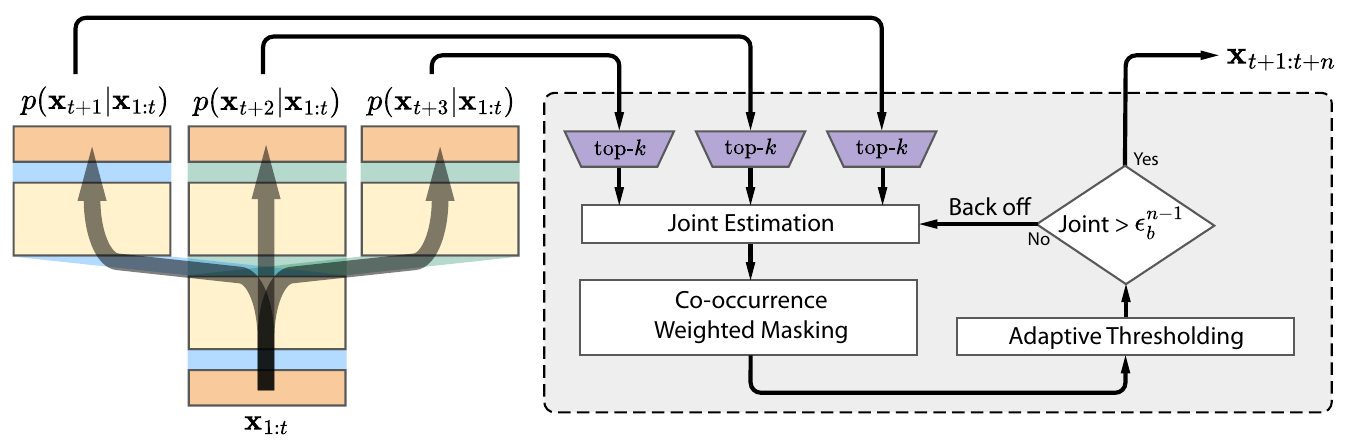}
    \caption{Flowchart of the proposed dynamic multi-token prediction pipeline.}
    \label{fig:flowchart}
\end{figure*}

Fig.~\ref{fig:flowchart} summarizes the proposed dynamic text generation pipeline. We extend the popular top-$k$ sampling
scheme~\citep{fan2018top_k, radford2019gpt2} for 
autoregressive language models to multi-token generation. 
First, we obtain logits for all token heads. We then obtain the top-$k$ probabilities for the predictions. Then, 
since we approximate the predicted tokens to be independent, we estimate the joint probability using 
Eq.~(\ref{eq:prob_joint_independent}). We bridge the gap between the true and the estimated (using independent predictions) joint probability 
distributions using co-occurrence weighted masking, taking inspiration from optimal transport~\citep{peyre2019optimal_transport}.
We fix the sparsity in higher-dimensional distributions using adaptive thresholding and backing off to lower-order $n$-gram prediction. 
We then sample from the joint probability distribution to output the generated sequence of tokens. Hence, DynaMo dynamically generates one 
or more tokens based on the given context and the model's confidence in its predictions. We describe the abovementioned methods next.

\subsubsection{Co-occurrence Weighted Masking}

To bridge the gap between the true and the estimated joint probability distribution in Eq.~(\ref{eq:prob_joint_independent}), we mask the
estimated distribution using the co-occurrence weights. Mathematically,
\begin{align}
\begin{split}
\label{eq:co_occurrence_approximaion_general}
    p(\rvx_{t+1:t+n} &| \rvx_{1:t}) \\
    &= \prod_{i=1}^n p(\rvx_{t+i} | \rvx_{1:t}) \frac{ p(\rvx_{t+1:t+n} | \rvx_{1:t}) }{ \prod_{i=1}^n p(\rvx_{t+i} | \rvx_{1:t}) } \\
    & \approx \prod_{i=1}^n f_\theta^i(\rvx_{1:t}) \underbrace{ \frac{ \hat{p}(\rvx_{t+1:t+n}) }{ \prod_{i=1}^n \hat{p}(\rvx_{t+i}) } }_{\text{co-occurrence mask}}
\end{split}
\end{align}
where $\hat{p}(\rvx_{t+1:t+n})$ and $\hat{p}(\rvx_{t+i})$ are sampled 
estimates of the joint probability and the prediction of the $i$-th token, respectively.
We estimate these probabilities based on the token counts in the training dataset. 
Note that the approximation in Eq.~(\ref{eq:co_occurrence_approximaion_general}) ignores the history $\rvx_{1:t}$. 

\newtheorem{theorem}{Theorem}
\begin{theorem}
When the cost function $c(\rvx_{t+1}, \rvx_{t+2}, \ldots, \rvx_{t+n}) = -\log \big( \frac{ \hat{p}(\rvx_{t+1:t+n}) }{ \prod_{i=1}^n \hat{p}(\rvx_{t+i}) } \big)$ and $\epsilon_2 = 0$ [defined in Eq.~(\ref{eq:optimal_transport})], the joint probability distribution in 
Eq.~(\ref{eq:co_occurrence_approximaion_general}) is the optimal solution to the optimal transport problem~\citep{peyre2019optimal_transport}.
\end{theorem}

We describe the optimal transport problem in the multi-token prediction setting and provide a proof of the above theorem 
in Appendix~\ref{app:optimal_transport}.

\subsubsection{Dynamic Back-off and Adaptive Thresholding}

Intuitively, when generating multiple tokens, the goal is to find the peaks in the predicted joint
probability distribution 
and sample those peaks. If none of the probability values is beyond a threshold (determined by $\epsilon_b$), i.e., there are no peaks in the joint probability distribution, our model backs off to lower-order $n$-gram prediction. To implement this, we adopt a static 
threshold $\epsilon_b$. If no probability value is $> \epsilon_b^{n-1}$, we back off to sampling a lower-order joint 
probability distribution. We set all probabilities less than $\epsilon_b$ to 0. 

Static thresholding is too na\"ive for joint probability distributions, 
which can vary with the predicted tokens and input context. Taking inspiration from computer vision methods, 
we test adaptive thresholding, leveraging Otsu's binarization algorithm~\citep{otsu1979threshold}. It adapts 
the threshold for dynamic back-off based on the predicted joint probability distribution. We apply adaptive thresholding on top of the static 
thresholding explained above. In other words, we first set all values in the joint probability distribution less 
than $\epsilon_b$ to 0. Then, we set all values less than $\epsilon_\text{AT}$ to 0 (where $\epsilon_\text{AT}$ 
is the threshold found using Otsu's algorithm). In the computer vision domain, researchers implement Otsu's 
algorithm after applying Gaussian blur to the input image. We thus explore the effect of using Gaussian blur 
and adaptive thresholding on the predicted joint probability distribution (ablation analysis in Appendix~\ref{app:ablation}). 

Alg.~\ref{alg:dynamic_multi-token_generation} summarizes the multi-token generation algorithm. We depict the probability 
distribution output by the $i$-th-token head by $f_\theta^i$. This probability distribution is a vector of 
length $V$ (or $k$ after top-$k$ sampling). We calculate the joint probability 
distribution $\mathbf{J}$ by taking the outer product of the individual token predictions. The function 
\texttt{adaptiveThresholding} (line~\ref{line:adaptive_thresholding}) implements adaptive thresholding explained above. The function \texttt{penalizeRepetition} (line~\ref{line:penalize_repetition}) divides all probabilities that correspond to repetitions by a penalty value~\citep{keskar2019ctrl}.
The \texttt{sample} function (lines~\ref{line:sample_joint} and \ref{line:sample_first_token_marginal}) samples the tokens using multinomial 
sampling, i.e., weighted by the corresponding probability values. Based on $n$, we output the sequence of 
generated tokens $\rvx_{t+1:t+n}$. For the proposed set of DynaMo models, we initialize $n = 3$. Thus, we dynamically 
generate new tokens depending on the output predictions (and the corresponding probabilities). A low value of 
$\epsilon_b$ generates more tokens (a three-token model with $\epsilon_b = 0$ will always generate three tokens). 
On the other hand, a high value of $\epsilon_b$ results in few tokens being generated ($\epsilon_b = 1$ will 
always generate only one token). 

\begin{algorithm}[t]
\caption{DynaMo multi-token generation} 
\label{alg:dynamic_multi-token_generation}
\begin{algorithmic}[1]
\Require input sequence $\rvx_{1:t}$, DynaMo model with token heads $f_\theta^i ,\ \forall \ i = 1, \ldots, n$.
\State $p(\rvx_{t+1}|\rvx_{1:t}) \gets f_\theta^1(\rvx_{1:t})$, 
\State $p(\rvx_{t+2}|\rvx_{1:t}) \gets f_\theta^2(\rvx_{1:t})$, 
\State $p(\rvx_{t+3}|\rvx_{1:t}) \gets f_\theta^3(\rvx_{1:t})$,
\State $n = 3$ (for three-token model)
\While{$n > 1$}
    \State Obtain top-$k$ values for token predictions $p(\rvx_{t+i}|\rvx_{1:t})$ 
    \State $\mathbf{J} \gets \prod_{i=1}^n f_\theta^i(\rvx_{1:t}) \frac{\hat{p}(\rvx_{t+1:t+n})}{\prod_{i=1}^n \hat{p}(\rvx_{t+i})}$ \\ 
    \Comment{Co-occurrence weighted masking}
    \State $\mathbf{J} \gets $ \texttt{adaptiveThresholding}($\mathbf{J}$) \label{line:adaptive_thresholding} \\ \Comment{Adaptive thresholding} 
    \State $\mathbf{J} \gets $ \texttt{penalizeRepetition}($\mathbf{J}$) \label{line:penalize_repetition} 
    \If{$j < \epsilon_b^{n-1}, \ \forall \ j  \in \mathbf{J}$}
        \State $n \gets n - 1$ \Comment{Back-off} 
    \Else
        \State $\rvx_{t+1:t+n} \gets $ \texttt{sample}($\mathbf{J}$) \label{line:sample_joint} 
        \State \textbf{return} $\rvx_{t+1:t+n}$
    \EndIf
\EndWhile
\State \textbf{return} $\rvx_{t+1:t+n} \gets $ \texttt{sample}($p(\rvx_{t+1}|\rvx_{1:t})$) \label{line:sample_first_token_marginal}
\end{algorithmic}
\end{algorithm}

\subsection{Evaluation Methods}

We propose various methods to evaluate our multi-token models. They include evaluating single-token prediction 
on standard natural language understanding (NLU) benchmarks, multi-token perplexity, and open-ended generation 
performance.

\subsubsection{NLU Benchmarks}

Evaluating multi-token prediction on NLU benchmarks is challenging. This is because most downstream benchmarks 
only require one-token prediction. However, we hypothesize that training a multi-token prediction transformer results in better prediction of even the first token. We call this a \emph{better} transformer. We evaluate our models on popular 
benchmarks with the first-token head. We use the \texttt{lm-evaluation-harness}~\citep{eval-harness} to carry out 
our evaluations on common benchmarks in both zero-shot and few-shot settings. For fair comparisons, we report the 
performance of the corresponding base Pythia model as well.

\subsubsection{Multi-token Perplexity}
\label{sec:multi_token_ppl}

To test multi-token text generation quality, we evaluate the models based on perplexity. However, the traditional 
definition of perplexity is only defined for single token prediction. We extend this to $n^{th}$ token prediction and 
also $n$-gram prediction. Mathematically,
\begin{align}
\begin{split}
    \text{PPL}_n &= \exp \left(\mkern-2mu -\frac{1}{T} \sum_{t=1}^{T-n} \log p(\rvx_{t+n} | \rvx_{1:t}) \mkern-2mu\right)\mkern-2mu, \\
    \text{PPL}_{1:n} &= \exp \left(\mkern-2mu -\frac{1}{nT} \sum_{t=1}^{T-n} \log p(\rvx_{t+1:t+n} | \rvx_{1:t}) \mkern-2mu\right)
\end{split}
\end{align}
For a three-token model, we calculate $\text{PPL}_1$, $\text{PPL}_{1:2}$, and $\text{PPL}_{1:3}$. We can also 
extend perplexity calculation to dynamic multi-token prediction, wherein we decide $n$ based on the joint 
probability distribution and the back-off threshold. We refer to it as $\text{PPL}_\text{d}$. It varies with $\epsilon_b$.

\begin{table*}[]
\centering
\caption{Zero-shot performance on common sense reasoning tasks.}
\resizebox{0.7\linewidth}{!}{
\begin{tabular}{@{}l|cccccccc@{}}
\toprule
\textbf{Model} & \textbf{ARC-c} & \textbf{ARC-e} & \textbf{BoolQ} & \textbf{COPA} & \textbf{HellaSwag} & \textbf{OBQA} & \textbf{PIQA} & \textbf{WinoG} \\ \midrule
Pythia-70M & 15.5$_{\pm 1.0}$ & 38.7$_{\pm 1.0}$ & \textbf{55.9$_{\pm 0.8}$} & 53.0$_{\pm 5.0}$ & 26.6$_{\pm 0.4}$ & 14.6$_{\pm 0.2}$ & 58.6$_{\pm 1.2}$ & \textbf{50.8$_{\pm 1.4}$} \\
DynaMo-77M-T3 & \textbf{17.3$_{\pm 1.1}$} & \textbf{41.0$_{\pm 1.0}$} & 55.7$_{\pm 0.9}$ & \textbf{56.0$_{\pm 5.0}$} & \textbf{26.9$_{\pm 0.4}$} & \textbf{14.7$_{\pm 1.6}$} & \textbf{59.8$_{\pm 1.1}$} & 49.8$_{\pm 1.4}$ \\ \midrule
Pythia-160M & \textbf{20.7$_{\pm 1.2}$} & 44.0$_{\pm 1.0}$ & \textbf{49.4$_{\pm 0.9}$} & 65.0$_{\pm 4.8}$ & 29.1$_{\pm 0.5}$ & \textbf{17.0$_{\pm 1.7}$} & 62.0$_{\pm 1.1}$ & 50.6$_{\pm 1.4}$ \\
DynaMo-180M-T3 & 19.4$_{\pm 1.1}$ & \textbf{45.3$_{\pm 1.0}$} & 48.0$_{\pm 0.9}$ & \textbf{66.0$_{\pm 4.8}$} & \textbf{29.3$_{\pm 0.5}$} & 16.6$_{\pm 1.7}$ & \textbf{62.7$_{\pm 1.1}$} & \textbf{51.7$_{\pm 1.4}$} \\ \midrule
Pythia-410M & 20.5$_{\pm 1.2}$ & 51.6$_{\pm 1.0}$ & \textbf{58.6$_{\pm 0.9}$} & \textbf{71.0$_{\pm 4.6}$} & 34.5$_{\pm 0.5}$ & 17.8$_{\pm 1.7}$ & 67.2$_{\pm 1.1}$ & \textbf{53.3$_{\pm 1.4}$} \\
DynaMo-430M-T3 & \textbf{21.2$_{\pm 1.2}$} & \textbf{52.6$_{\pm 1.0}$} & 57.1$_{\pm 0.9}$ & 70.0$_{\pm 4.6}$ & \textbf{34.6$_{\pm 0.5}$} & \textbf{17.9$_{\pm 1.7}$} & \textbf{67.5$_{\pm 1.1}$} & \textbf{53.3$_{\pm 1.4}$} \\ \midrule
Pythia-1B & 24.3$_{\pm 1.2}$ & \textbf{58.5$_{\pm 1.0}$} & 60.8$_{\pm 0.9}$ & 74.0$_{\pm 4.4}$ & \textbf{38.9$_{\pm 0.5}$} & 21.8$_{\pm 1.8}$ & 70.1$_{\pm 1.1}$ & 52.9$_{\pm 1.4}$ \\
DynaMo-1.1B-T3 & \textbf{25.3$_{\pm 1.3}$} & 58.4$_{\pm 1.0}$ & \textbf{60.9$_{\pm 0.9}$} & \textbf{76.0$_{\pm 4.3}$} & \textbf{38.9$_{\pm 0.5}$} & \textbf{22.2$_{\pm 1.9}$} & \textbf{70.2$_{\pm 1.1}$} & \textbf{53.8$_{\pm 1.4}$} \\ \midrule
Pythia-1.4B & 27.3$_{\pm 1.3}$ & \textbf{61.8$_{\pm 1.0}$} & 58.0$_{\pm 0.9}$ & 76.0$_{\pm 4.3}$ & 41.7$_{\pm 0.5}$ & \textbf{22.8$_{\pm 1.9}$} & 72.0$_{\pm 1.0}$ & \textbf{56.9$_{\pm 1.4}$} \\
DynaMo-1.5B-T3 & \textbf{27.7$_{\pm 1.3}$} & 61.5$_{\pm 1.0}$ & \textbf{59.2$_{\pm 0.9}$} & \textbf{78.0$_{\pm 4.2}$} & \textbf{41.9$_{\pm 0.5}$} & 22.4$_{\pm 1.9}$ & \textbf{72.5$_{\pm 1.0}$} & 56.0$_{\pm 1.4}$ \\ \midrule
Pythia-2.8B & 29.9$_{\pm 1.3}$ & 53.5$_{\pm 1.0}$ & \textbf{64.2$_{\pm 0.8}$} & 75.0$_{\pm 4.4}$ & 45.4$_{\pm 0.5}$ & 24.0$_{\pm 1.9}$ & 74.1$_{\pm 1.0}$ & 58.2$_{\pm 1.4}$ \\
DynaMo-2.9B-T3 & \textbf{30.4$_{\pm 1.3}$} & \textbf{64.7$_{\pm 1.0}$} & 64.0$_{\pm 0.8}$ & \textbf{80.0$_{\pm 4.0}$} & \textbf{45.7$_{\pm 0.5}$} & \textbf{24.3$_{\pm 1.9}$} & \textbf{74.2$_{\pm 1.0}$} & \textbf{59.1$_{\pm 1.4}$} \\ \midrule
Pythia-6.9B & 33.2$_{\pm 1.4}$ & \textbf{68.5$_{\pm 1.0}$} & 64.4$_{\pm 0.8}$ & 74.0$_{\pm 4.4}$ & 49.6$_{\pm 0.5}$ & 27.0$_{\pm 1.9}$ & \textbf{75.7$_{\pm 1.0}$} & 62.7$_{\pm 1.4}$ \\
DynaMo-7.3B-T3 & \textbf{33.6$_{\pm 1.4}$} & 68.1$_{\pm 1.0}$ & \textbf{65.1$_{\pm 0.8}$} & \textbf{76.0$_{\pm 4.3}$} & \textbf{49.9$_{\pm 0.5}$} & \textbf{28.0$_{\pm 2.0}$} & \textbf{75.7$_{\pm 1.0}$} & \textbf{62.9$_{\pm 1.4}$} \\ \bottomrule
\end{tabular}}
\label{tbl:commonsense}
\end{table*}

\subsection{Open-ended Text Generation}
\label{sec:open_ended_generation}

Perplexity is a very restrictive evaluation measure. It constrains model text generation to the text in the 
validation set. A fairer approach to test multi-token generation would be to evaluate open-ended generated
texts. \citet{zheng2023llm_as_a_judge} propose using strong LLMs like GPT-3.5~\citep{chatgpt} and GPT-4~\citep{openai2023gpt4} and show that they can 
match both controlled and crowdsourced human preferences in evaluating generated texts well. Since human 
evaluation of open-ended generated texts from our models would be very expensive and time-consuming, we use a 
strong LLM to evaluate the quality of generated text from our DynaMo suite of models. 

Vicuna and MT benchmarks~\citep{zheng2023llm_as_a_judge} require the pre-trained LLM to be finetuned on 
instruction-following datasets. To disambiguate the effect of instruction finetuning, we evaluate our models 
with different target speed-ups on a novel \emph{sentence-completion} benchmark. The task is to complete a 
sentence for a given prompt. We categorize the sentences into simple declarative, compound declarative, 
W/H interrogative, Y/N interrogative, affirmative imperative, negative imperative, and exclamatory. We test
the text generations of our models for grammatical correctness, creativity, depth, logical flow, coherence, 
and informativeness of the generated text. The benchmark has ten prompts. For every prompt, we generate ten 
sentences with different random seeds for every $\epsilon_b \in \{0.00, 0.02, \ldots, 1.00\}$. Thus, for every 
model, we generate 5100 sentences at different speed-ups. We evaluate the quality of every generated sentence 
using single-mode and pairwise evaluations. For single-mode evaluation, we ask GPT-3.5 to score the generated 
response from one to ten. For pairwise evaluation, we ask GPT-3.5 to compare the response against one generated 
by the corresponding Pythia base model. DynaMo either wins, loses, or ties against the baseline Pythia model. 
We provide further details on the sentence completion benchmark along with the evaluation setup in 
Appendix~\ref{app:sentence_completion_bench}.

Finally, we also evaluate the performance of instruction-finetuned DynaMo models on the Vicuna benchmark. We use 
the Alpaca dataset~\citep{alpaca} filtered by GPT-3.5 for high-quality instruction-response 
pairs~\citep{chen2023alpagasus}. The dataset contains 9,229 instruction-response pairs. We follow the evaluation 
setup from \cite{zheng2023llm_as_a_judge}.
\section{Experiments}
\label{sec:results}

In this section, we present experimental results and comparisons of the proposed approach with the Pythia baseline, 
which we used to instantiate the DynaMo models. We provide test results for architectural and training variations
in multi-token prediction in Appendix~\ref{app:multi_token_prediction_methods}.

\subsection{Downstream Performance}
\label{sec:downstream_perf}

We hypothesize that training the decoder layers using the 
second- and third-token loss terms makes them \emph{better}. We test this hypothesis next.

We consider eight standard common sense reasoning benchmarks: ARC challenge (ARC-c) and ARC easy 
(ARC-e,~\citealt{clark2018arc}), 
BoolQ~\citep{clark2019boolq}, COPA~\citep{roemmele2011copa}, HellaSwag~\citep{zellers2019hellaswag}, OpenBookQA 
(OBQA,~\citealt{mihaylov2018obqa}), PIQA~\citep{bisk2020piqa}, and WinoGrande (WinoG,~\citealt{sakaguchi2021winogrande}). 
We perform evaluations in the zero-shot setting as done in the language modeling community. Table~\ref{tbl:commonsense} shows
a comparison between each model in the DynaMo suite with that of the corresponding baseline Pythia model. 
As we can see, DynaMo models outperform their respective baselines on most benchmarks. We report additional
downstream performance results in Appendix~\ref{app:downstream_perf_additional}.

\subsection{Multi-token Perplexity}

Table~\ref{tbl:multi_token_ppl} shows the multi-token perplexity on 
the validation set for all models in the DynaMo and Pythia suites. The DynaMo models achieve lower $\text{PPL}_1$ 
relative to their Pythia counterparts due to further training of the first-token head and \emph{better} decoder layers in the model stem (i.e., all layers up to the penultimate layer). We provide further test results in 
Appendix~\ref{app:multi_token_prediction_methods}. The multi-token perplexity drops as models 
become larger, making the prediction of multiple tokens easier and better. We describe results for dynamic multi-token perplexity
($\text{PPL}_\text{d}$) in Appendix~\ref{app:dynamic_multi_token_ppl}.

\begin{table}[t!]
\centering
\caption{Multi-token perplexity results for models in the DynaMo and Pythia suites.}
\resizebox{\linewidth}{!}{
\begin{tabular}{@{}l|ccccc@{}}
\toprule
\textbf{Model} & $\mathbf{PPL_1}$  & $\mathbf{PPL_2}$   & $\mathbf{PPL_3}$   & $\mathbf{PPL_{1:2}}$ & $\mathbf{PPL_{1:3}}$ \\ \midrule
Pythia-70M     & 20.2$_{\pm  1.5}$ & -                  & -                  & -                   & -                    \\
DynaMo-77M-T3  & \textbf{18.3$_{\pm  1.5}$} & 111.4$_{\pm  1.7}$ & 262.0$_{\pm  1.6}$ & 45.2$_{\pm  1.5}$   & 81.2$_{\pm  1.6}$    \\ \midrule
Pythia-160M    & 13.5$_{\pm 1.4}$  & -                  & -                  & -                   & -                    \\
DynaMo-180M-T3 & \textbf{12.9$_{\pm 1.4}$}  & 78.5$_{\pm 1.6}$   & 199.4$_{\pm 1.6}$  & 31.8$_{\pm 1.5}$    & 58.7$_{\pm 1.5}$     \\ \midrule
Pythia-410M    & 9.9$_{\pm 1.4}$   & -                  & -                  & -                   & -                    \\
DynaMo-430M-T3 & \textbf{9.6$_{\pm 1.4}$}   & 59.8$_{\pm 1.6}$   & 162.4$_{\pm 1.6}$  & 24.0$_{\pm 1.5}$    & 45.4$_{\pm 1.5}$     \\ \midrule
Pythia-1B      & 8.5$_{\pm 1.4}$   & -                  & -                  & -                   & -                    \\
DynaMo-1.1B-T3 & \textbf{8.4$_{\pm 1.4}$}   & 44.1$_{\pm 1.6}$   & 116.6$_{\pm 1.7}$  & 19.3$_{\pm 1.5}$    & 35.1$_{\pm 1.6}$     \\ \midrule
Pythia-1.4B    & 7.9$_{\pm 1.6}$   & -                  & -                  & -                   & -                    \\
DynaMo-1.5B-T3 & \textbf{7.8$_{\pm 1.6}$}   & 41.9$_{\pm 2.0}$   & 112.7$_{\pm 2.1}$  & 18.3$_{\pm 1.9}$    & 33.6$_{\pm 1.9}$     \\ \midrule
Pythia-2.8B    & 7.4$_{\pm 1.6}$   & -                  & -                  & -                   & -                    \\
DynaMo-2.9B-T3 & \textbf{7.1$_{\pm 1.9}$}   & 37.1$_{\pm 2.7}$   & 100.3$_{\pm 3.0}$  & 16.2$_{\pm 2.2}$    & 29.8$_{\pm 2.4}$     \\ \midrule
Pythia-6.9B    & 6.6$_{\pm 1.8}$   & -                  & -                  & -                   & -                    \\
DynaMo-7.3B-T3   & \textbf{6.5$_{\pm 1.8}$}   &  31.4$_{\pm 2.6}$  &  83.5$_{\pm 3.0}$  &  14.4$_{\pm 2.2}$   &  25.8$_{\pm 2.4}$   \\ \bottomrule
\end{tabular}}
\label{tbl:multi_token_ppl}
\end{table}

\subsection{Text Generation Performance and Speed-up}

We now compare the open-ended text generation performance of the DynaMo models with that of the baseline Pythia models 
on the sentence-completion benchmark.

Since pairwise evaluations by strong LLMs better align with human evaluations~\citep{zheng2023llm_as_a_judge}, we 
evaluate our models against the Pythia baseline in the pairwise mode (details in 
Appendix~\ref{app:sentence_completion_bench}; single-mode evaluations in Appendix~\ref{app:single_mode_evaluation}). As $\epsilon_b$ increases, the text quality 
improves, but the speed-up decreases. Thus, the win rate (i.e., the number of wins/losses against the baseline) 
decreases as speed-up increases. 

\begin{figure}
    \centering
    \includegraphics[width=\linewidth]{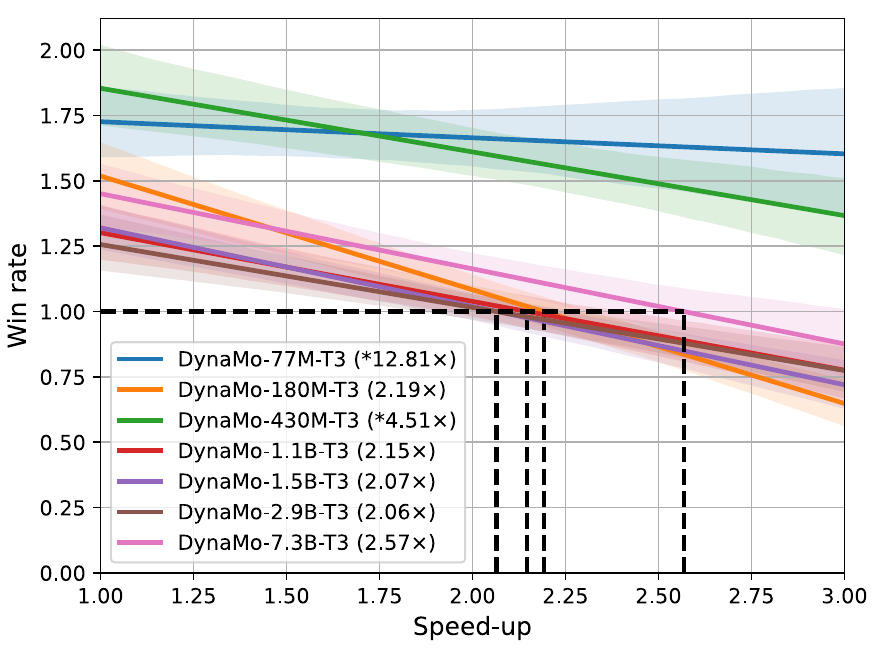}
    \caption{Win rate vs.~speed-up for pairwise comparisons on the sentence-completion benchmark with
corresponding Pythia models as baselines. GPT-3.5 is used as a judge. Regression plotted with 95\% confidence intervals. Same-quality speed-ups are shown in parentheses. Theoretical same-quality speed-ups are marked with an asterisk (*).}
    \label{fig:win_rate_vs_speedup}
\end{figure}

Fig.~\ref{fig:win_rate_vs_speedup} shows the effect of speed-up on the win rate of the proposed models (we describe how we obtain
this plot in Appendix~\ref{app:pairwise_evaluation}). When the 
win rate is 1.0, the text generation quality would, on average, be the same for the models being compared. We 
call the speed-up for this case the ``same-quality speed-up.'' If the win rate for a model is always greater than 
1.0, we extrapolate the plot to obtain the ``\emph{theoretical} same-quality speed-up.'' However, in further 
discussions, we refer to the minimum of (theoretical) same-quality speed-up and 3$\times$ (for three-token models) 
as, simply, the ``speed-up.''

\subsection{Instruction Finetuning}

\begin{figure*}[t!]
    \centering
    \includegraphics[width=0.9\linewidth]{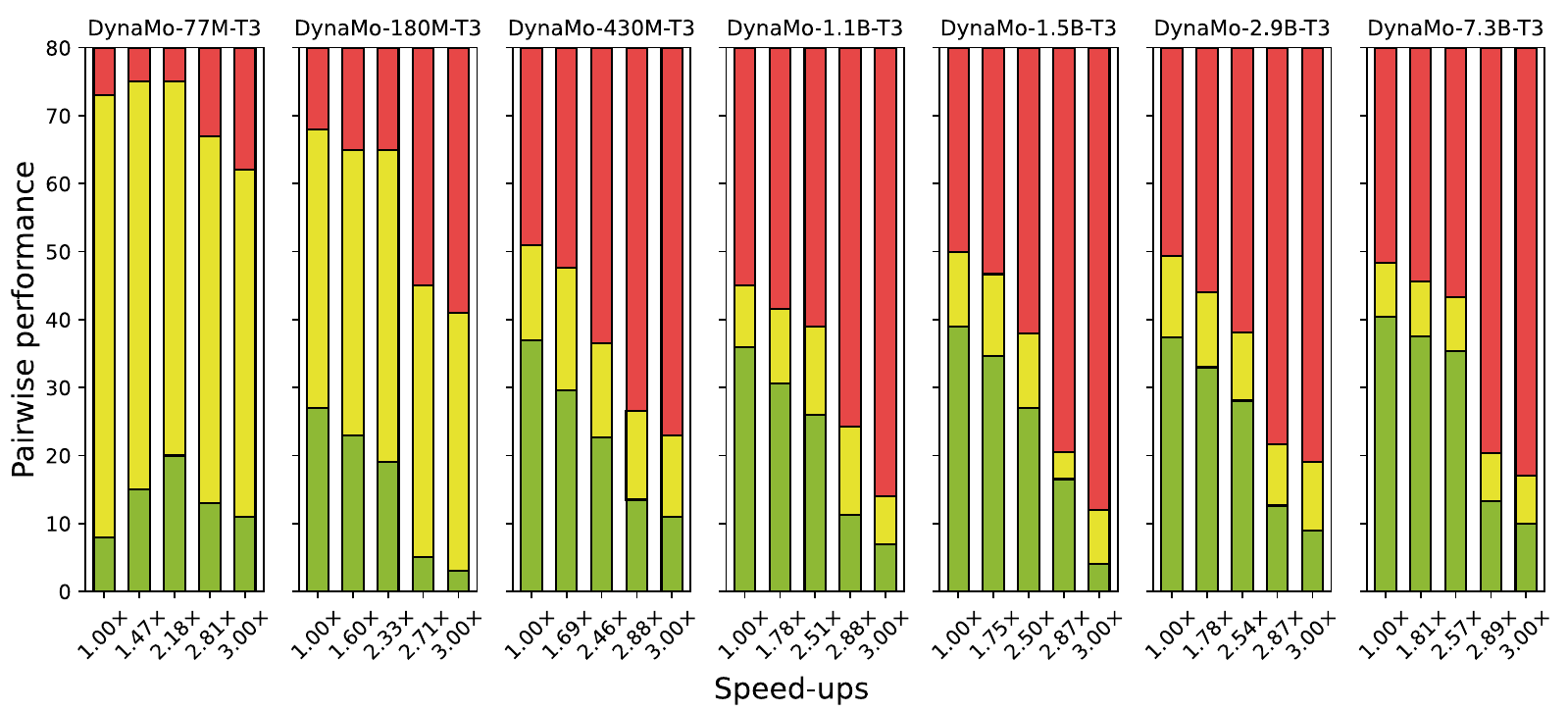}
    \caption{Pairwise performance of the DynaMo and Pythia models on the Vicuna benchmark. GPT-4 was used as a 
judge. The actual number of wins, ties, and losses are colored green, yellow, and red, respectively.}
    \label{fig:pairwise_mode_vicuna}
\end{figure*}

We finetune models in the Pythia and DynaMo suites on an instruction-following dataset (details in 
Section~\ref{sec:open_ended_generation}). Fig.~\ref{fig:pairwise_mode_vicuna} shows the pairwise performance 
of the DynaMo (with respect to Pythia) models on the Vicuna benchmark~\citep{zheng2023llm_as_a_judge}. We run 
the DynaMo models at different speed-ups (we set $\epsilon_b = 1.0, 0.75, 0.5, 0.25, 0.0$) shown on the $x$-axis. 
We compare each model against the corresponding Pythia baseline. In the case of comparisons with small models, 
neither model results in a reasonable answer. Hence, GPT-4 classifies many response pairs as ties. The number of 
ties decreases as model sizes increase. As the speed-up increases, the win rate decreases. DynaMo-7.3B-T3 provides 
around the same-quality responses as Pythia-6.9B (win rate = 0.98) even for a high speed-up of 2.57$\times$ (we ablate the
effect of dynamic text generation methods in Appendix~\ref{app:ablation}).
\section{Discussion}
\label{sec:discussion}

In this section, we discuss the implications of the proposed DynaMo suite of multi-token prediction models and 
future work directions. 

\begin{table*}[]
\centering
\caption{Effect of \emph{better} transformer training on zero-shot performance in common sense tasks.}
\resizebox{0.7\linewidth}{!}{
\begin{tabular}{@{}l|cccccccc@{}}
\toprule
\textbf{Model} & \textbf{ARC-c} & \textbf{ARC-e} & \textbf{BoolQ} & \textbf{COPA} & \textbf{HellaSwag} & \textbf{OBQA} & \textbf{PIQA} & \textbf{WinoG} \\ \midrule
Pythia-70M & 15.5$_{\pm 1.0}$ & 38.7$_{\pm 1.0}$ & \textbf{55.9$_{\pm 0.8}$} & 53.0$_{\pm 5.0}$ & 26.6$_{\pm 0.4}$ & 14.6$_{\pm 0.2}$ & 58.6$_{\pm 1.2}$ & 50.8$_{\pm 1.4}$ \\
Pythia-70M$^+$ & 15.6$_{\pm 1.0}$ & 38.8$_{\pm 1.0}$ & \textbf{55.9$_{\pm 0.8}$} & 53.1$_{\pm 5.0}$ & 26.8$_{\pm 0.4}$ & 14.6$_{\pm 0.2}$ & 58.6$_{\pm 1.2}$ & \textbf{50.9$_{\pm 1.4}$} \\ \midrule
DynaMo-77M-T3 & \textbf{17.3$_{\pm 1.1}$} & \textbf{41.0$_{\pm 1.0}$} & 55.7$_{\pm 0.9}$ & \textbf{56.0$_{\pm 5.0}$} & \textbf{26.9$_{\pm 0.4}$} & \textbf{14.7$_{\pm 1.6}$} & \textbf{59.8$_{\pm 1.1}$} & 49.8$_{\pm 1.4}$ \\ \bottomrule
\end{tabular}}
\label{tbl:commonsense_stronger}
\end{table*}

\subsection{Effect of Better Transformer Training}
\label{app:stronger_attention_heads}

Another observation that supports the hypothesis that \emph{better} transformer training results in superior first-token prediction is as follows. For fair comparisons, we test our three-token model against Pythia-70M further trained on the 5\% Pile dataset using a learning rate of $10^{-5}$ 
(we refer to this version as Pythia-70M$^+$) on commonsense tasks. We present the result in Table~\ref{tbl:commonsense_stronger} (perplexity results in Appendix~\ref{app:ppl_evaluations}). Training the decoder layers based on the modified-CLM loss in Eq.~(\ref{eq:clm_multi}) results in better first-token prediction, which we use to evaluate common sense tasks as presented here. This key result is worth further exploration, which we leave for future work.

\subsection{Contribution of Unigram, Bigram, and Trigram Generations to Speed-up}

\begin{figure}[t!]
    \centering
    \includegraphics[width=\linewidth]{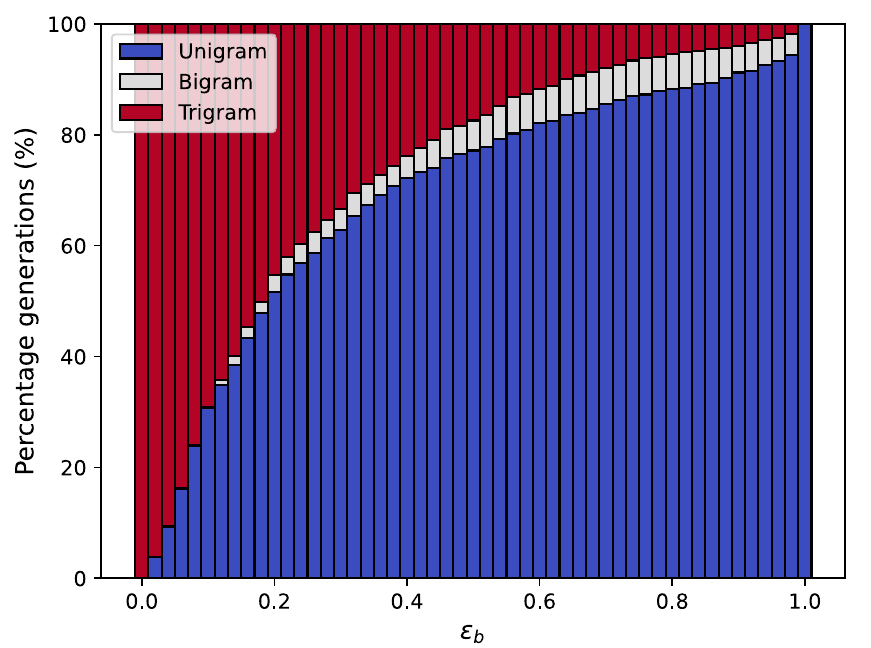}
    \caption{Percentage of unigram, bigram, and trigram generations vs. $\epsilon_b$ for DynaMo-70M-T3.}
    \label{perc_gen}
\end{figure}

Fig.~\ref{perc_gen} shows the percentage of one-token, two-token, and three-token generations as we sweep $\epsilon_b$ with DynaMo-70M-T3. When $\epsilon_b = 1.0$, the model always generates one token at a time. When $\epsilon_b = 0.0$, the model always generates three tokens at a time, regardless of its confidence in the generations. Surprisingly, we note that the contribution of two-token generations is low; the model banks on three-token generations instead. We defer further exploration to balance multi-token generations during dynamic back-off to future work.

\subsection{Baseline Comparisons}

Table~\ref{tbl:related} shows comparisons with other approaches that target inference speed-up. Speculative 
sampling~\citep{chen2023speculative} and skeleton-of-thought decoding~\citep{ning2023skeleton} are orthogonal 
to the DynaMo approach and can be used in conjunction with the proposed multi-token generation scheme to boost 
performance further. Nevertheless, DynaMo can be seen to require the least overhead in FLOPS-per-generation 
and provide the highest speed-up. The high computational efficiency of DynaMo is attributed to its avoidance of
high-batch operations necessitated by speculative sampling and skeleton-of-thought decoding.

\begin{table}[t!]
\centering
\caption{Comparisons with other approaches. $^*$\citet{ning2023skeleton} evaluate models of different sizes.}
\resizebox{\linewidth}{!}{
\begin{tabular}{@{}l|ccc@{}}
\toprule
\textbf{Method} & \textbf{Base Model Size} & \textbf{FLOPS Overhead} & \textbf{Speed-up} \\ \midrule
Speculative Sampling & 70B & 340\% & 1.92-2.46$\times$ \\ \midrule
Skeleton-of-Thought & 7B-13B$^*$ & 560\% & 1.13-2.39$\times$ \\ \midrule
RecycleGPT & 1.3B & 15\% & 1.34-1.40$\times$ \\ \midrule
DynaMo-77M-T3 & 70M & 8.95\% & \textbf{3.00$\times$} \\
DynaMo-180M-T3 & 160M & 8.73\% & 2.19$\times$ \\
DynaMo-430M-T3 & 410M & 6.22\% & \textbf{3.00$\times$} \\
DynaMo-1.1B-T3 & 1B & 9.95\% & 2.15$\times$ \\
DynaMo-1.5B-T3 & 1.4B & 7.12\% & 2.07$\times$ \\
DynaMo-2.5B-T3 & 2.4B & \textbf{5.67\%} & 2.06$\times$ \\
DynaMo-7.3B-T3 & 6.9B & 5.87\% & 2.57$\times$ \\ \bottomrule
\end{tabular}}
\label{tbl:related}
\end{table}

\subsection{How Many Tokens Can We Simultaneously Predict?}

\begin{figure}[t!]
    \centering
    \includegraphics[width=\linewidth]{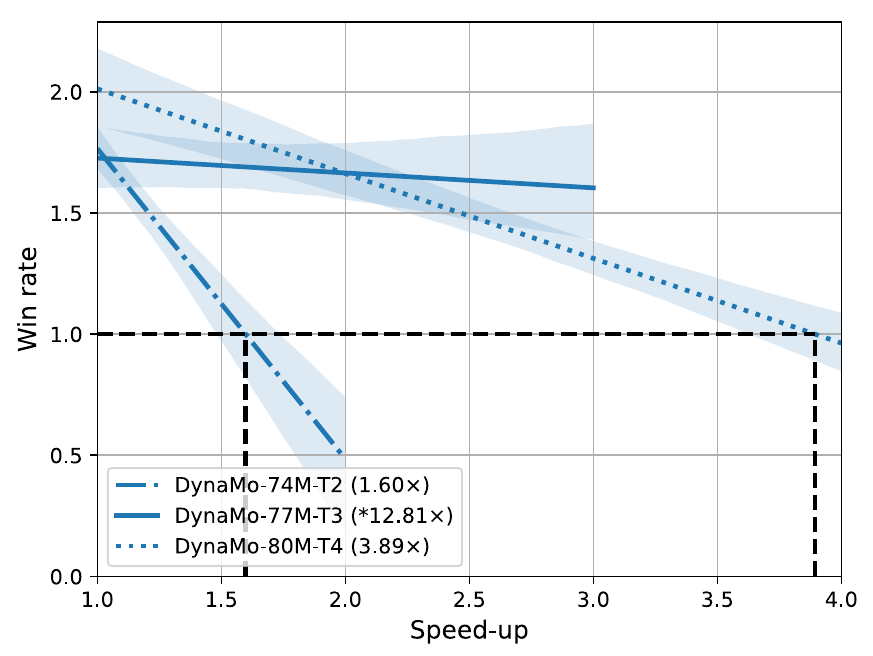}
    \caption{Win rate vs.~speed-up for pairwise comparisons on the sentence-completion benchmark with Pythia-70M 
as the baseline. GPT-3.5 is used as a judge. Theoretical same-quality speed-up is marked with an asterisk (*).}
    \label{fig:win_rate_vs_speedup_t2_vs_t3_vs_t4}
\end{figure}

Fig.~\ref{fig:win_rate_vs_speedup_t2_vs_t3_vs_t4} shows the win rates with respect to speed-ups on the sentence-completion 
benchmark using pairwise analysis against Pythia-70M (see Section~\ref{sec:open_ended_generation} and Appendix~\ref{app:sentence_completion_bench}). 
DynaMo-77M-T3 shows much better win rates relative to DynaMo-74M-T2 for speed-ups $<$ 2.0 despite similar 
$\text{PPL}_{1:2}$. Further, DynaMo-77M-T3, being a three-token model, can provide much higher speed-ups than 
DynaMo-74M-T2, however, at the cost of a slight parameter overhead. Since the extra parameter overhead is 
marginal, especially for larger models, we stick with three-token models.

We also explore simultaneous token prediction beyond the three-token model. Fig.~\ref{fig:win_rate_vs_speedup_t2_vs_t3_vs_t4} also shows the performance of DynaMo-80M-T4. Due to the \emph{better} transformer training through the modified-CLM objective, the four-token model achieves higher win rates than the three-token counterpart for speed-ups $<$ 2.0. DynaMo-80M-T4 achieves a same-quality speed-up of 3.89$\times$, however, at an additional parameter overhead. Apart from the parameter overhead, the quadruplet co-occurrence mask incurs additional memory overhead. While the pairwise and triplet masks (calculated over 5\% of the Pile dataset) only occupy 53.43MB and 152.59MB, respectively, the quadruplet mask (calculated over 0.05\% of the Pile dataset) occupies 3.33GB memory. We store all co-occurrence masks using the sparse coordinate format. This overhead may still be negligible for very large models (>7B parameters). We leave simultaneous prediction of more than four tokens and optimized implementation of corresponding co-occurrence masks to future work.

\subsection{Additional Benchmarking}

We show the performance of the DynaMo models on most downstream benchmarking tasks. These results show that the
\emph{better} transformer trained using loss terms for predicting subsequent tokens generally leads to 
improved downstream performance while incurring no significant adverse effect on the model's bias and 
misinformation abilities (see Appendix~\ref{sec:bias_and_misinformation}). While \citet{mukherjee2023orca} suggest evaluating world knowledge acquisition through tasks like  AGIEval~\citep{zhong2023agieval} and Big-Bench Hard~\citep{suzgun2022bbh}, we defer assessing larger multi-token models on such complex benchmarks to future work.

\section{Conclusion}
\label{sec:conclusion}

In this work, we presented DynaMo, a suite of multi-token prediction language models. We trained the proposed 
model suite efficiently by reusing weights of existing pre-trained LLMs. We proposed novel ways to dynamically 
predict multiple tokens for a given context. The DynaMo models dynamically back off to lower-order $n$-gram 
prediction based on a threshold. We also proposed adaptive thresholding and co-occurrence weighted masking on 
the modeled joint probability distribution to improve text generation quality. One of our proposed models, DynaMo-7.3B-T3, 
achieved the same-quality generated text as the baseline (Pythia-6.9B) while achieving 2.57$\times$ speed-up 
with only 5.87\% and 2.67\% parameter and training time overheads (see Appendix~\ref{app:training_overheads}).
\section{Limitations}
\label{sec:limitation}

We trained DynaMo models on only 5\% of the Pile dataset~\citep{gao2020pile}. However, training the models on the 
entire dataset would further boost performance due to improved estimates of the joint probability distributions. 
Future multi-token models can directly be trained on the entire language corpus without the complex multi-learning-rate 
learning employed here (details in Appendix~\ref{app:training_and_evaluation_process}). Finally, the current suite of DynaMo models was trained with the Pythia backbone. One could also leverage state-of-the-art open-source foundation models~\citep{touvron2023llama2} to train the DynaMo suite.
\section*{Acknowledgments}
\label{sec:ack}

This work was supported by Samsung Research America, Mountain View. N. K. J. was supported by NSF under Grant No. CCF-2203399.

\bibliography{biblio}

\clearpage
\begin{appendices}

\section{Experimental Setup Details}
\label{sec:experimental_setup_details}

In this section, we provide details on the training and evaluation processes along with other hyperparameters. 
We then describe the sentence-completion benchmark. Finally, we present the overheads in training time for our
DynaMo suite of models.

\subsection{Training and Evaluation Processes}
\label{app:training_and_evaluation_process}

To train the DynaMo suite of models, we first transfer the weights from the base Pythia model. Then, we train the 
models on a randomly sampled 5\% set of 
sentences in the Pile dataset\footnote{Dataset source: \url{https://huggingface.co/datasets/EleutherAI/pile-deduped-pythia-random-sampled}.}. We train for one epoch on this dataset. We choose a subset of the same dataset on which the base 
Pythia model was trained to avoid catastrophic forgetting when being trained on a different dataset. In the future, 
we plan to train the models on other datasets using standard continual learning approaches~\citep{de2021continual}.

\begin{figure}[t!]
    \centering
    \includegraphics[width=0.8\linewidth]{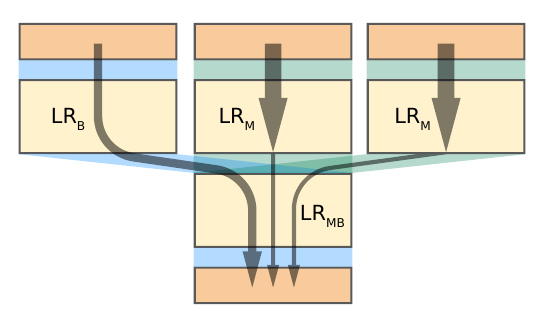}
    \caption{Gradient flow when training a DynaMo model.}
    \label{fig:multi_token_learning_rates}
\end{figure}

We now describe the training procedure for the DynaMo suite of models. First, we transfer the weights for the base model
(i.e., the model stem and the final decoder layer). Then, we train the base model with a 
low learning rate ($\text{LR}_{\text{B}}$). On the other hand, we train subsequent token heads using a higher 
learning rate ($\text{LR}_{\text{M}}$) since we randomly initialize their weights. However, when backpropagating 
those gradients to the model stem, we use a much lower learning rate ($\text{LR}_{\text{MB}}$). We hypothesize that 
when the decoder layers learn from the first and subsequent token predictions, they make the transformer \emph{better} 
in predicting multiple tokens.
Table~\ref{tbl:learning_rates} shows the learning rates used for different models in the DynaMo suite.
Fig.~\ref{fig:multi_token_learning_rates} shows the gradient flow when training an example three-token DynaMo 
model. 

\begin{figure}[t!]
    \centering
    \includegraphics[width=\linewidth]{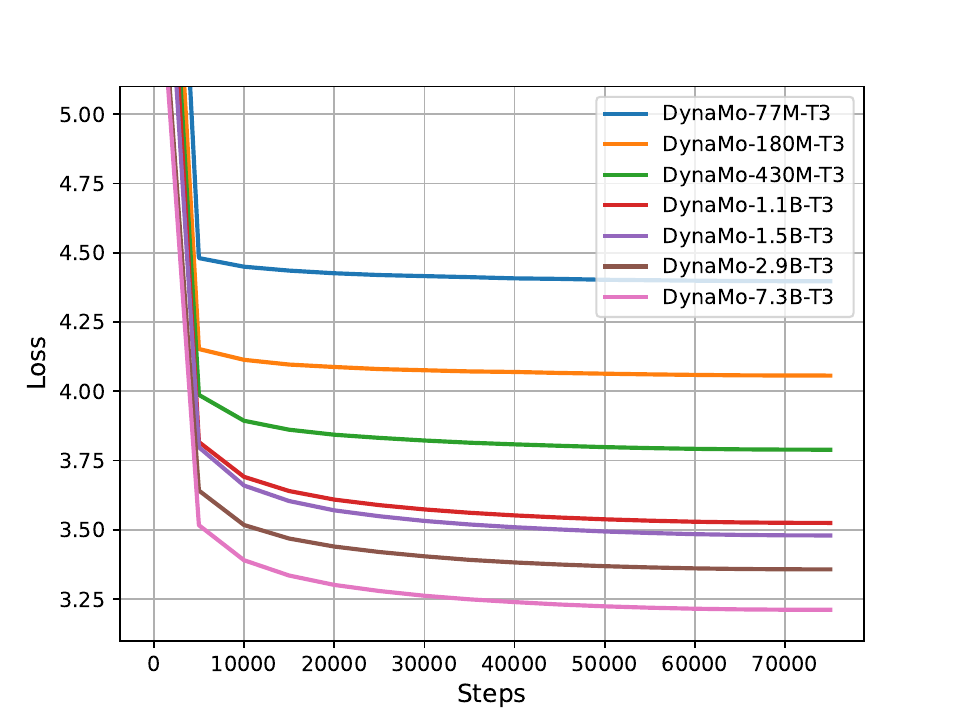}
    \caption{Loss curves for three-token models in the DynaMo suite.}
    \label{fig:loss_all_models}
\end{figure}

We train our models using the AdamW optimizer~\citep{loshchilov2017adamw} with the following hyperparameters: 
$\beta_1 = 0.9, \beta_2 = 0.95, \epsilon = 1 \times 10^{-8}$. We use the cosine learning rate scheduler such that 
the learning rate warms up for 1\% of the dataset (758 steps) and then drops to 0 at the end of training. We use a 
batch size of 64 sentences, i.e., 131,072 tokens (each sentence is 2,048 tokens long). The dataset has 5M sentences, 
which we divide into a training set (97\%) and validation set (3\%). Thus, a batch size of 64 results in 75,782 
training steps in one training epoch. We evaluate the model at every 5,000 steps. Fig.~\ref{fig:loss_all_models} 
shows the three-token validation loss (logarithm of $\text{PPL}_{1:3}$) for models in the DynaMo suite.

\begin{table}[]
\centering
\caption{Learning rates used for training different models in the DynaMo suite.}
\resizebox{0.8\linewidth}{!}{
\begin{tabular}{@{}l|ccc@{}}
\toprule
\textbf{Model} & $\mathbf{LR_{B}}$ & $\mathbf{LR_{M}}$ & $\mathbf{LR_{MB}}$ \\ \midrule
DynaMo-77M-T3 & $10^{-5}$ & $10^{-3}$ & $10^{-6}$ \\
DynaMo-180M-T3 & $6 \times 10^{-6}$ & $6 \times 10^{-4}$ & $6 \times 10^{-7}$ \\
DynaMo-430M-T3 & $3 \times 10^{-6}$ & $3 \times 10^{-4}$ & $3 \times 10^{-7}$ \\
DynaMo-1.1B-T3 & $2 \times 10^{-6}$ & $2 \times 10^{-4}$ & $2 \times 10^{-7}$ \\
DynaMo-1.5B-T3 & $2 \times 10^{-6}$ & $2 \times 10^{-4}$ & $2 \times 10^{-7}$ \\
DynaMo-2.9B-T3 & $1.6 \times 10^{-6}$ & $1.6 \times 10^{-4}$ & $1.6 \times 10^{-7}$ \\
DynaMo-7.3B-T3 & $1.2 \times 10^{-6}$ & $1.2 \times 10^{-4}$ & $1.2 \times 10^{-7}$ \\ \bottomrule
\end{tabular}}
\label{tbl:learning_rates}
\end{table}

We train the models on A100 GPUs with 80GB memory. For efficient implementation of our models, we use the 
\texttt{flash-attention} library~\citep{dao2022flashattention}. Our models also support memory-efficient attention 
in the \texttt{xformers} library~\citep{xFormers2022}. Since DynaMo-7.3B-T3 did not fit in memory, we resorted to 
PyTorch's fully-sharded data parallel (FSDP) training feature. Table~\ref{tbl:fsdp} provides the hyperparameters used 
for the FSDP configuration.

\begin{table}[]
\centering
\caption{FSDP configuration used for training DynaMo-7.3B-T3.}
\resizebox{0.8\linewidth}{!}{
\begin{tabular}{@{}ll@{}}
\toprule
\textbf{Configuration Key} & \textbf{Value} \\ \midrule
Sharding strategy & \texttt{SHARD\_GRAD\_OP} \\
Transformer-based wrap & \texttt{DYNAMO\_LAYER} \\
All-gather backward prefetch policy & \texttt{BACKWARD\_PRE} \\
All-gather forward prefetch policy & \texttt{NONE} \\
Mixed precision & \texttt{FP16} \\ \bottomrule
\end{tabular}}
\label{tbl:fsdp}
\end{table}

For text generation, we use $k = 50$ for top-$k$ decoding, temperature $= 0.7$, and repetition penalty $= 1.1$. 
The default text generation hyperparameters for the DynaMo models are $\alpha_c = 1.0$ (see Appendix~\ref{app:ablation}), adaptive thresholding with 
Gaussian blur (kernel size $= 3$), and using co-occurrence weighted masking unless otherwise specified.

\subsection{Training Overheads}
\label{app:training_overheads}

Table~\ref{tbl:training_overheads} shows the overhead of training models in the DynaMo suite. We report training 
times for modified-CLM training on 5\% of the Pile dataset and instruction-finetuning. We present the reported 
CLM training times for the Pythia models~\citep{biderman2023pythia}. 

\begin{table}[t!]
\centering
\caption{Training (with overheads) and instruction-finetuning times for the DynaMo suite of models.}
\resizebox{\linewidth}{!}{
\begin{tabular}{@{}l|cc@{}}
\toprule
\textbf{Model} & \textbf{Training GPU Hrs.} & \textbf{Instruction-FT GPU Mins.} \\ \midrule
Pythia-70M & 510 & - \\
DynaMo-77M-T3 & 15 (2.94\%) & 8 \\ \midrule
Pythia-160M & 1,030 & - \\
DynaMo-180M-T3 & 36 (3.49\%) & 15 \\ \midrule
Pythia-410M & 2,540 & - \\
DynaMo-430M-T3 & 46 (1.81\%) & 30 \\ \midrule
Pythia-1B & 4,830 & - \\
DynaMo-1.1B-T3 & 80 (1.65\%) & 60 \\ \midrule
Pythia-1.4B & 7,120 & - \\
DynaMo-1.5B-T3 & 88 (1.24\%) & 72 \\ \midrule
Pythia-2.8B & 14,240 & - \\
DynaMo-2.9B-T3 & 176 (1.24\%) & 180 \\ \midrule
Pythia-6.9B & 33,500 & - \\
DynaMo-7.3B-T3 & 896 (2.67\%) & 864 \\ \bottomrule
\end{tabular}}
\label{tbl:training_overheads}
\end{table}

\subsection{Sentence-completion Benchmark}
\label{app:sentence_completion_bench}

In this section, we provide details of the sentence-completion benchmark. This benchmark is motivated by the 
Vicuna benchmark~\citep{zheng2023llm_as_a_judge}. However, it is meant for pre-trained LLMs that are not 
instruction-finetuned. This dissociates any effects of instruction-finetuning from model performance. The 
benchmark consists of ten prompts requiring the model to complete the sentence. These prompts correspond to 
sentences of different types. Table~\ref{tbl:sentence_completion_prompts} outlines the prompts.

\begin{table}[t!]
\centering
\caption{Prompts in the sentence-completion benchmark.}
\resizebox{0.9\linewidth}{!}{
\begin{tabular}{@{}ll@{}}
\toprule
\textbf{Prompt} & \textbf{Type} \\ \midrule
\texttt{I am a student at the} & Simple Declarative \\
\texttt{This is going to be a very} & Simple Declarative \\
\texttt{He wanted to play, but} & Compound Declarative \\
\texttt{How can we} & W/H Interrogative \\
\texttt{What will} & W/H Interrogative \\
\texttt{Will you} & Y/N Interrogative \\
\texttt{Please explain} & Affirmative Imperative \\
\texttt{Do not} & Negative Imperative \\
\texttt{Wow! I can't believe that} & Exclamatory \\
\texttt{This is amazing! We} & Exclamatory \\ \bottomrule
\end{tabular}}
\label{tbl:sentence_completion_prompts}
\end{table}

To obtain the GPT score, we ask GPT-3.5 to rate the generated sentence on a scale from 1 to 10. For pairwise 
evaluations, we ask GPT-3.5 to compare the generated text (by our DynaMo model) against a baseline (the 
corresponding baseline Pythia model) and rate it as a ``win,'' ``lose,'' or a ``tie.'' We use 
\texttt{gpt-3.5-turbo-0613} for our evaluations. Fig.~\ref{fig:single_mode_prompt} shows the prompt template 
used for single-mode evaluations and Fig.~\ref{fig:pairwise_mode_prompt} shows the prompt template used for 
pairwise evaluations.  However, this benchmark also suffers from the same drawbacks as the Vicuna 
benchmark~\citep{zheng2023llm_as_a_judge}, which we attempt to alleviate. To address position bias in pairwise 
comparisons, we randomly order the responses of the assistants.

\begin{figure}[t!]
\centering
\resizebox{0.8\linewidth}{!}{
\begin{tcolorbox}
    \texttt{Please act as an impartial judge and evaluate the quality of the response provided by an AI assistant to the input prompt. The AI assistant provides an open-ended generation for the input prompt. Your evaluation should be based on the grammatical correctness, creativity, depth, logical flow, coherence, and based on how informative the response is. Do not let the length of the generated text influence your evaluation. Be as objective as possible. Begin your evaluation by providing a short explanation. Explain the mistakes, if any. After providing your explanation, you must rate the response on a scale of 1 to 10 by strictly following this format: "[[rating]]", for example: "Rating: [[5]]"}
\end{tcolorbox}}
\caption{Prompt template to rate the sentence quality of the candidate assistant model on an absolute scale 
(single-mode evaluation).}
\label{fig:single_mode_prompt}
\end{figure}

\begin{figure}[t!]
\centering
\resizebox{0.8\linewidth}{!}{
\begin{tcolorbox}
    \texttt{Please act as an impartial judge and evaluate the quality of the responses provided by two AI assistants to the input prompt. Both AI assistants provide open-ended generations for the input prompt. You should choose the assistant that produces a better generation. Your evaluation should be based on the grammatical correctness, creativity, depth, logical flow, coherence, and based on how informative the responses are. Do not let the lengths of the generated texts influence your evaluation. Do not favor certain names of the assistants. Begin your evaluation by comparing the two responses and provide a short explanation. Explain the mistakes, if any. Avoid any positional biases and ensure that the order in which the responses were presented does not influence your decision. Be as objective as possible. After providing your explanation, output your final verdict by strictly following this format: "[[A]]" if assistant A is better, "[[B]]" if assistant B is better, and "[[C]]" for a tie.}
\end{tcolorbox}}
\caption{Prompt template to rate the sentence quality of the candidate assistant model against a baseline model (pairwise-mode evaluation).}
\label{fig:pairwise_mode_prompt}
\end{figure}

\section{Optimal Transport Theory}
\label{app:optimal_transport}

Eq.~(\ref{eq:prob_joint_independent}) approximates the output joint probability by directly multiplying the 
independent marginal distributions. This implicitly assumes that $\rvx_{t+2}$ is independent of $\rvx_{t+1}$ 
conditioned on history $\rvx_{1:t}$, $\rvx_{t+3}$ is independent of $\rvx_{t+1}$ and 
$\rvx_{t+2}$, and so on. The downside of this decoding strategy is that it ignores the fact that the prediction of 
$\rvx_{t+2}$ depends heavily on which $\rvx_{t+1}$ is chosen (and similarly for subsequent predictions). A simple example is to consider 
$\rvx_{1:t} =$ \texttt{I}; here, \texttt{to} is a plausible second-word prediction as many sentences lead to 
that word, such as \texttt{I like to}, \texttt{I want to}, and \texttt{I went to}. On the other hand, 
\texttt{am} is a plausible first-word prediction. However, as long as one chooses it, the weight for \texttt{to} 
as the second-word prediction should be minimal unless we want to make our English teacher cry. This motivates 
us to weight the joint probability distribution based on co-occurrence of words (or, more precisely, tokens).

What follows is a theoretical motivation behind the use of co-occurrence weighted masking. Formally, according
to optimal transport theory~\citep{peyre2019optimal_transport}, we define a cost function $c(\rvx_{t+1},
\ldots, \rvx_{t+n}), \ \forall \ \rvx_{t+1}, \ldots, \rvx_{t+n}$. Once we define the 
cost function, we pose the joint estimation problem as follows,
\begin{align}
\label{eq:optimal_transport}
\begin{split}
    \arg \min_p &\sum p(\rvx_{t+1:t+n} | \rvx_{1:t}) c(\rvx_{t+1}, \ldots, \rvx_{t+n}) \\
    &\Delta \rvx_{t+1} \ldots \Delta \rvx_{t+n} \\
    +~&\epsilon_1 \text{KL}\left(p(\rvx_{t+1:t+n} | \rvx_{1:t}) || \prod_{i=1}^n f_\theta^i(\rvx_{1:t}) \right) \\
    +~& \epsilon_2 \sum_{i=1}^n \text{KL}\left(p(\rvx_{t+i} | \rvx_{1:t}) || f_\theta^i(\rvx_{1:t}) \right)
\end{split}
\end{align}
Although solving an optimal transport problem is fast, using the celebrated Sinkhorn 
algorithm~\citep{sejourne2019sinkhorn}, we propose the use of Eq.~(\ref{eq:co_occurrence_approximaion_general}) as an 
approximation that works well in practice, as we demonstrate in our experimental results. Next, we show that 
the approximation in Eq.~(\ref{eq:co_occurrence_approximaion_general}) is indeed the closest to preserving the true joint 
probability distribution, when the correction term (co-occurrence mask) is not dependent on the history $\rvx_{1:t}$.

\begin{figure*}[t]
    \centering
    \includegraphics[width=0.9\linewidth]{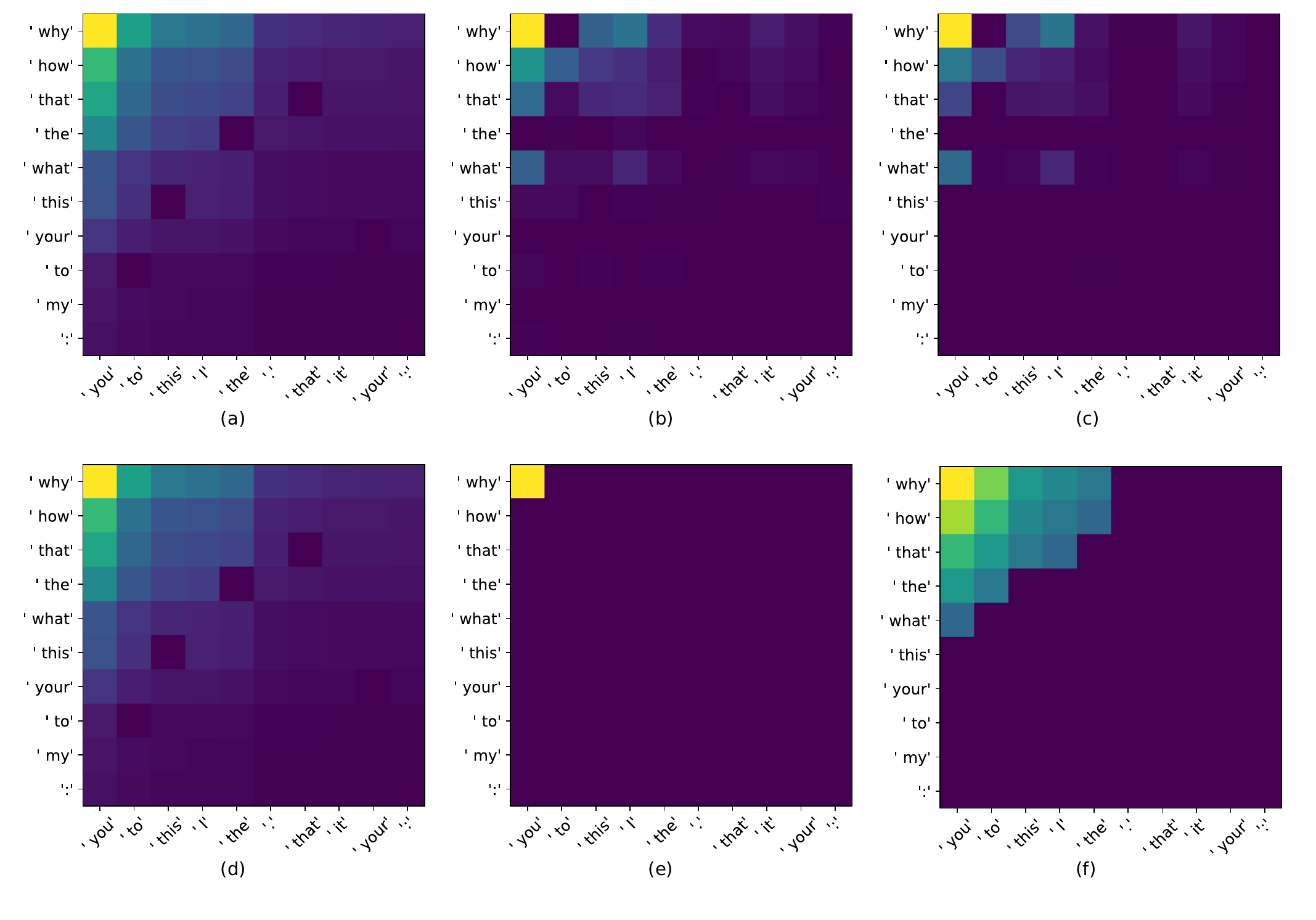}
    \caption{Joint probability distribution with top 10 tokens sorted in decreasing order of probabilities using the 
DynaMo-2.9B-T2 model for the input prompt: \texttt{Please explain}. Probabilities corresponding to repetition 
have been penalized by a factor of 100. (a) and (d) are vanilla distributions. Co-occurrence masked distribution 
with (b) $\alpha_c = 0.5$ [CO-0.5] and (c) $\alpha_c = 1.0$ [CO]. Adaptive thresholding (e) without Gaussian 
blur [AT], and (f) with Gaussian blur (kernel size = 3) [AT + G-3].}
    \label{fig:joint_combined}
\end{figure*}

\begin{proof}[Proof of Theorem 1.]
Recall that the optimization in Eq.~(\ref{eq:optimal_transport}) is subject to the constraint $\int p(\rvx_{t+1:t+n} | \rvx_{1:t}) \Delta \rvx_{t+1} \ldots \Delta \rvx_{t+n} = 1$. Thus, the Lagrangian of the objective is given by
\begin{align*}
    L = &\sum p(\rvx_{t+1:t+n} | \rvx_{1:t}) c(\rvx_{t+1}, \ldots, \rvx_{t+n}) \\
    &\Delta \rvx_{t+1} \ldots \Delta \rvx_{t+n} \\
    &+ \epsilon_1 \text{KL}\left(p(\rvx_{t+1:t+n} | \rvx_{1:t}) || \prod_{i=1}^n f_\theta^i(\rvx_{1:t}) \right) \\
    &+ \epsilon_2 \sum_{i=1}^n \text{KL}\left(p(\rvx_{t+i} | \rvx_{1:t}) || f_\theta^i(\rvx_{1:t}) \right) \\
    &+ \lambda \Big(\sum p(\rvx_{t+1:t+n} | \rvx_{1:t}) \Delta \rvx_{t+1} \ldots \Delta \rvx_{t+n} \\
    &- 1 \Big)
\end{align*}
Setting the derivative of $L$ w.r.t. $p(\rvx_{t+1:t+n} | \rvx_{1:t})$ to zero, we get
\begin{align*}
    p^*(&\rvx_{t+1:t+n} | \rvx_{1:t}) \\
    &\propto \prod_{i=1}^n f_\theta^i(\rvx_{1:t}) \ \exp \left( c(\rvx_{t+1}, \ldots, \rvx_{t+n})/\epsilon_1 \right) \\
    &= \prod_{i=1}^n f_\theta^i(\rvx_{1:t}) \frac{\hat{p}(\rvx_{t+1:t+n})}{\prod_{i=1}^n \hat{p}(\rvx_{t+i})}
\end{align*}
\end{proof}

\section{Additional Results}
\label{app:additional_results}

In this section, we report additional supporting results.

\subsection{Ablation of Dynamic Text Generation Methods}
\label{app:ablation}

In this section, we ablate the effect of adaptive thresholding (with and without Gaussian blur) and co-occurrence 
weighted masking (see Section~\ref{sec:dynamic_text_generation}). Figs.~\ref{fig:joint_combined}(a)-(c) show 
the effect of co-occurrence masking on the two-token joint probability with decreasing masking transparency $\alpha_c$. 
Mathematically, we modify Eq.~(\ref{eq:co_occurrence_approximaion_general}) for the two-token prediction case as follows:
\begin{align}
\begin{split}
\label{eq:co_occurrence_alpha}
    p(\rvx_{t+1}&, \rvx_{t+2} | \rvx_{1:t}) \\
    &\approx f_\theta^1(\rvx_{1:t}) f_\theta^2(\rvx_{1:t}) \left( \frac{\hat{p}(\rvx_{t+1}, \rvx_{t+2})}{\hat{p}(\rvx_{t+1}) \ \hat{p}(\rvx_{t+2})} \right)^{\alpha_c} 
\end{split}
\end{align}
where $\alpha_c = 1.0$ implies that the co-occurrence weights mask the joint probability distribution with no 
transparency. On the other hand, we do not use co-occurrence masking when $\alpha_c = 0.0$. Nevertheless, 
$\alpha_c = 0.5$ partially masks the joint probability distribution using the co-occurrence weights. 
Figs.~\ref{fig:joint_combined}(d)-(f) show the effect of adaptive thresholding with and without Gaussian blur.

\begin{figure}[t!]
    \centering
    \includegraphics[width=\linewidth]{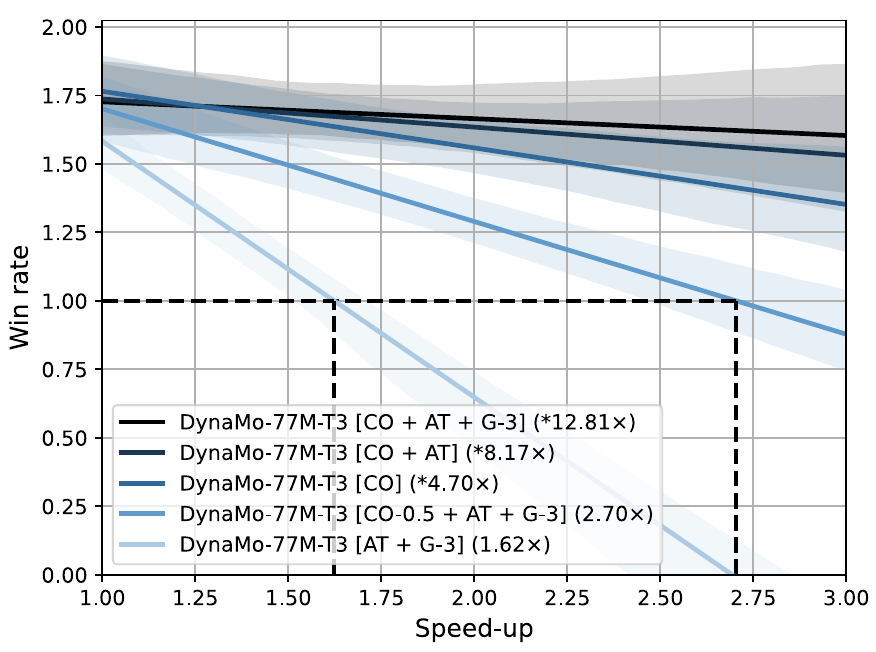}
    \caption{Ablation analysis using adaptive thresholding (with and without Gaussian blur) and co-occurrence 
masking. Win rates for pairwise tests against Pythia-70M on the sentence-completion benchmark are shown for 
different speed-ups. GPT-3.5 is used as the judge. Theoretical same-quality speed-ups are marked with an 
asterisk (*).}
    \label{fig:win_rate_vs_speedup_ablation}
\end{figure}

\begin{figure*}[t!]
    \centering
    \includegraphics[width=0.8\linewidth]{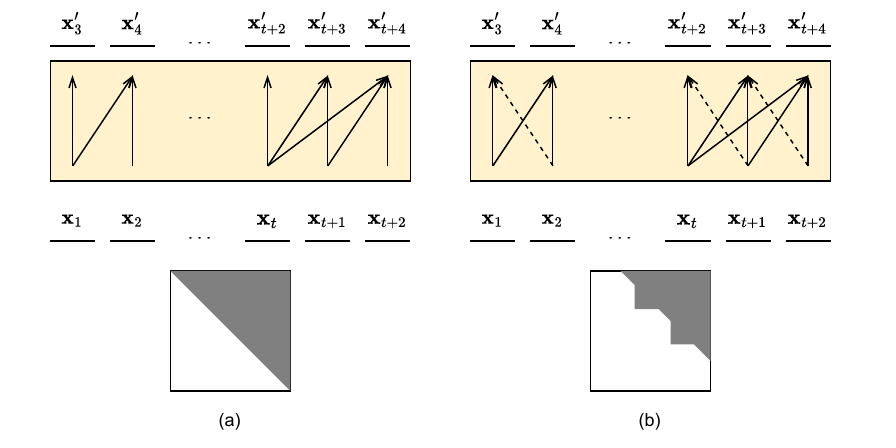}
    \caption{Multi-token prediction using a single-token head. The input sequence is shown below the transformer layer. The model 
predicts the output sequence above. Attention arrows correspond to the modified CLM objective. The attention masks 
are shown below the input sequences. (a) T1-L2-M0: labels are shifted by two positions (i.e., the model predicts $\rvx'_{t+2}$ with 
$\rvx_t$ as input). Under the modified CLM objective, the model learns to predict $\rvx'_{t+2} = \rvx_{t+2}$. 
(b) T1-L2-M(-1)R: labels are shifted by two positions but masks are shifted in the opposite direction (i.e., for 
predicting $\rvx'_{t+2}$, the model can sometimes see $\rvx_{t+1}$).}
    \label{fig:label_mask_shift}
\end{figure*}

Fig.~\ref{fig:win_rate_vs_speedup_ablation} shows the win rates vs.~speed-up for DynaMo-77M-T3, where we generated 
the texts in the sentence-completion benchmark using difference schemes. We observe that co-occurrence masking 
(with $\alpha_c = 1.0$, i.e., the default setting used in our experiments) used along with adaptive thresholding 
(after application of Gaussian blur with a kernel size $=$ 3) results in the flattest win rate vs.~speed-up 
curve, thus, providing the highest theoretical same-quality speed-up.

We ablate the effect of dynamic text generation methods with the instruction-finetuned DynaMo-7.3B-T3 model on the Vicuna benchmark in Table~\ref{tbl:win_rate_vicuna_ablation}. We take the case $\epsilon_b = 0.5$ (that results in 2.57$\times$ speed-up in Fig.~\ref{fig:pairwise_mode_vicuna}) and present the win rates against Pythia-6.9B. Leveraging co-occurrence weighted masking along with adaptive thresholding using Gaussian blur (kernel size $=$ 3) results in the highest win rate.

\begin{table}[t]
\centering
\caption{Ablations analysis of dynamic text generation methods with the instruction-finetuned DynaMo-7.3B-T3 model on the Vicuna benchmark. We use $\epsilon_b = 0.5$.}
\resizebox{0.6\linewidth}{!}{
\begin{tabular}{@{}l|cc@{}}
\toprule
\textbf{Method} & \textbf{Speed-up} & \textbf{Win rate}  \\ \midrule
CO + AT + G-3 & 2.57$\times$ & \textbf{0.98} \\
CO + AT & 2.44$\times$ & 0.96 \\
CO & \textbf{2.61$\times$} & 0.82 \\
CO-0.5 + AT + G3 & 2.55$\times$ & 0.77 \\
AT + G-3 & 2.49$\times$ & 0.38 \\ \bottomrule
\end{tabular}}
\label{tbl:win_rate_vicuna_ablation}
\end{table}

\subsection{Exploration of Multi-token Prediction Methods}
\label{app:multi_token_prediction_methods}

In this section, we provide a detailed overview of various architectural and training variations tested for 
multi-token prediction.

\subsubsection{Design Variations}

Under the CLM objective, the attention mask prevents the model 
from seeing future tokens, i.e., we only compute the attentions corresponding to the lower triangular matrix 
(we refer to this case as M0). In summary, we represent traditional autoregressive models 
as T1-L1-M0. We study different variations of the above formulation for multi-token prediction. These include multiple token 
heads, label shifts, and mask shifts. We explore them below. After testing various approaches, we observe that for, say, 
three-token prediction, the T3-L1-M0 set of choices performs the best. Thus, in all discussions in the main paper, we 
represent DynaMo-T3-L1-M0 as simply DynaMo-T3. 

Fig.~\ref{fig:label_mask_shift} shows the information 
flow for T1-L2-M0 and T1-L2-M(-1)R cases. In the former case, for predicting $\rvx_{t+2}$, the model only sees 
the input context $\rvx_{1:t}$. Hence, we shift the mask in the latter case. However, T1-L2-M(-1) would be 
equivalent to the traditional T1-L1-M0 (ignoring residual connections that result in information leakage). Hence, 
we randomly mask out some tokens so that the model learns to predict the next and the second-next token at each 
position. Another position-equivalent modeling approach to T1-L2-M(-1)R is T1-L1-M1R. However, both these modeling 
approaches suffer from information leakage. T1-L2-M(-1)R suffers from information leakage due to expanding 
receptive fields along model depth. We fix this by incorporating negative mask shifts only in the first layer of 
the LLM. T1-L1-M1R suffers from information leakage due to the residual/skip connections in the LLM. Hence, we do 
not use this approach and test T1-L2-M(-1)R instead.

\begin{figure}[t!]
    \centering
    \includegraphics[width=\linewidth]{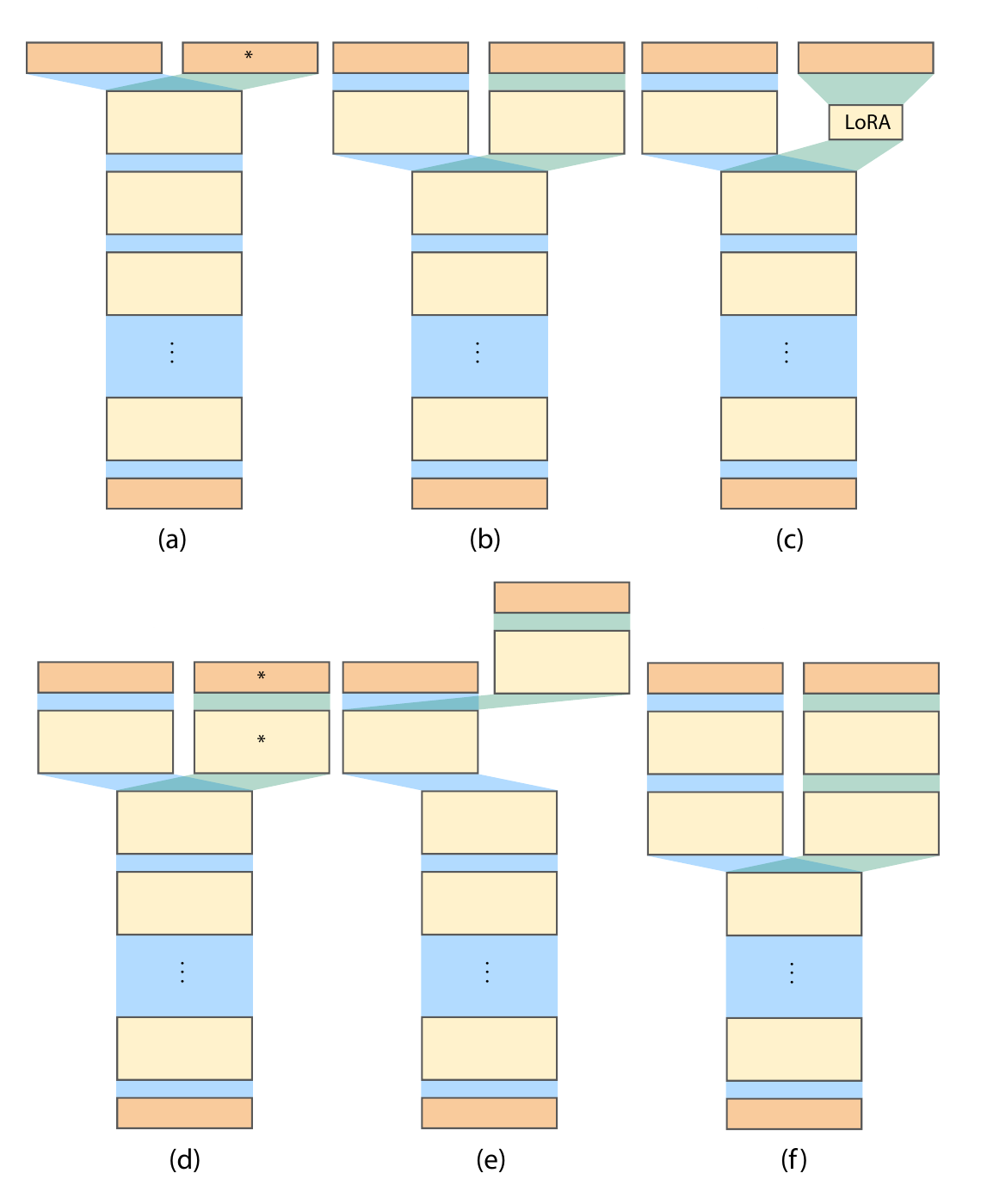}
    \caption{Architectural variations of the two-token prediction model that we tested: (a) DynaMo-96M-T2, 
(b) DynaMo-74M-T2 (C), (c) DynaMo-70M-T2 (LoRA), (d) DynaMo-99M-T2, (e) DynaMo-74M-T2 (NP), and (f) DynaMo-77M-T2.}
    \label{fig:arch_variations}
\end{figure}

Fig.~\ref{fig:arch_variations} shows different architectural variations of the two-token model we tested. We 
initialize all these models from the base Pythia-70M model. Fig.~\ref{fig:arch_variations}(a) shows the schematic 
of DynaMo-96M-T2 that randomly initializes the output embedding for the second-token head (we denote newly 
initialized weights by $\ast$ while other variations reuse these weights). The output embedding has 26M trainable 
parameters. Fig.~\ref{fig:arch_variations}(b) shows DynaMo-74M-T2 (C), which copies the weights of the decoder 
layer for the second-token head from the last layer of the first-token head (or the base model). Its output 
embedding for the second-token head reuses the weights from the first-token head. Since we copy the weights, we 
train the copied weights with a low learning rate ($\text{LR}_\text{B}$). Fig.~\ref{fig:arch_variations}(c) shows 
DynaMo-70M-T2 (LoRA) with only 65K trainable parameters~\citep{hu2021lora}. The LoRA module includes a low-rank 
matrix (we use rank $=$ 32). We add its output to that of the last decoder layer for second-token prediction. 
Fig.~\ref{fig:arch_variations}(d) shows DynaMo-99M-T2. We train a decoder layer and the output embedding for the 
second-token head, where we randomly initialize the weights of both modules. Fig.~\ref{fig:arch_variations}(e) 
shows DynaMo-74M-T2 (NP), where we feed the output of the last layer of the base model to the decoder layer for 
the second-token head. All models in the DynaMo suite use the outputs of the penultimate layer of the base model 
for subsequent token prediction. Instead, this model uses the output of the final (non-penultimate or NP) layer. 
Finally, Fig.~\ref{fig:arch_variations}(f) shows the use of two decoder layers for the second-token head.

\subsubsection{Evaluations}
\label{app:ppl_evaluations}

\begin{table}[t]
\centering
\caption{Multi-token perplexity results for various architectural variations. $^+$Model was further trained on 5\% Pile dataset.}
\label{tbl:ppl_arch_variations}
\resizebox{\linewidth}{!}{
\begin{tabular}{@{}l|ccccc@{}}
\toprule
\textbf{Model} & $\mathbf{PPL_1}$ & $\mathbf{PPL_2}$ & $\mathbf{PPL_3}$ & $\mathbf{PPL_{12}}$ & $\mathbf{PPL_{123}}$ \\ \midrule
Pythia-70M & 20.2$_{\pm 1.5}$ & - & - & - & - \\ 
Pythia-70M$^+$ & 20.1$_{\pm 1.5}$ & - & - & - & - \\ \midrule
DynaMo-70M-T1-L2 & 21.4$_{\pm 1.6}$ & 1455.8$_{\pm 6.4}$ & - & 189.3$_{\pm 2.2}$ & - \\
DynaMo-70M-T1-L2-M(-1)R & 20.3$_{\pm 1.5}$ & 645.3$_{\pm 1.9}$ & - & 87.4$_{\pm 1.7}$ & - \\ \midrule
DynaMo-96M-T2 & 19.9$_{\pm 1.5}$ & 252.4$_{\pm 1.9}$ & - & 68.0$_{\pm 1.5}$ & - \\ \midrule
DynaMo-74M-T2 (C) & 18.3$_{\pm 1.5}$ & 296.4$_{\pm 1.5}$ & - & 73.7$_{\pm 1.5}$ & - \\
DynaMo-70M-T2 (LoRA) & 20.2$_{\pm 1.5}$ & 1368.1$_{\pm 1.8}$ & - & 161.2$_{\pm 1.6}$ & - \\ \midrule
DynaMo-74M-T2 (CTC) & 18.5$_{\pm 1.5}$ & 115.4$_{\pm 1.7}$ & - & 46.0$_{\pm 1.6}$ & - \\ \midrule
DynaMo-99M-T2 & 18.3$_{\pm 1.5}$ & 111.5$_{\pm 1.7}$ & - & 45.2$_{\pm 1.5}$ & - \\
DynaMo-74M-T2 (NP) & 18.8$_{\pm 1.5}$ & 131.1$_{\pm 1.6}$ & - & 49.0$_{\pm 1.5}$ & - \\
DynaMo-74M-T2-H & 20.2$_{\pm 1.5}$ & 119.1$_{\pm 1.7}$ & - & 49.0$_{\pm 1.5}$ & - \\
DynaMo-74M-T2 & 18.3$_{\pm 1.5}$ & 112.4$_{\pm 1.7}$ & - & 45.4$_{\pm 1.5}$ & - \\
DynaMo-77M-T2 & 18.3$_{\pm 1.5}$ & 86.7$_{\pm 1.7}$ & - & 39.9$_{\pm 1.6}$ & - \\ \midrule
DynaMo-77M-T3 & 18.3$_{\pm 1.5}$ & 111.4$_{\pm 1.7}$ & 262.0$_{\pm 1.6}$ & 45.2$_{\pm 1.5}$ & 81.2$_{\pm 1.6}$ \\ \bottomrule
\end{tabular}}
\end{table}

Table~\ref{tbl:ppl_arch_variations} shows the multi-token perplexity results for various architectural and 
training variations of the DynaMo model with Pythia-70M as the baseline. For fair comparisons, we also add the
perplexity results for Pythia-70M$^+$ (trained using $\text{LR}_{\text{B}} = 10^{-5}$). It does not result in a lower $\text{PPL}_1$. This shows that with traditional CLM training, $\text{PPL}_1$ has converged. However, with the modified-CLM training (details in Appendix~\ref{app:training_and_evaluation_process}), $\text{PPL}_1$ for models in the DynaMo suite goes down further.
The architectural variations are as 
explained above. DynaMo-74M-T2 (CTC) shows the perplexity results for the model trained using CTC 
loss~\citep{yan2023ctc}. DynaMo-74M-T2-H is the model where we only train the decoder layer of the second-token 
head. Training this model is much faster than training DynaMo-74M-T2, as we need to calculate only a few 
gradients. However, this does not make the decoder layers in the model stem \emph{better}. We see that 
$\text{PPL}_1$ of this model is the same as that of Pythia-70M. One could increase the parameter budget for 
multi-token prediction by either adding another decoder layer for predicting the second token (DynaMo-77M-T2) 
or using a decoder layer for the third-token head (DynaMo-77M-T3). In the DynaMo suite of models, we traded the 
parameter budget for higher speed-up (using three-token models). We leave the exploration and search among 
various architectural decisions~\citep{chitty2022neural, tuli2023edgetran} targeting text generation performance 
and speed-up to future work.

\subsection{Additional Downstream Performance Results}
\label{app:downstream_perf_additional}

We now present additional results on downstream benchmarks.

\begin{table}[t!]
\centering
\parbox{0.4\linewidth}{
\centering
\caption{Five-shot exact match performance on the TriviaQA benchmark.}
\resizebox{1.0\linewidth}{!}{
\begin{tabular}{@{}l|c@{}}
\toprule
\textbf{Model} & \textbf{TriviaQA} \\ \midrule
Pythia-70M & \textbf{0.2$_{\pm 0.0}$} \\
DynaMo-77M-T3 & \textbf{0.2$_{\pm 0.0}$} \\ \midrule
Pythia-160M & 2.1$_{\pm 0.1}$ \\
DynaMo-180M-T3 & \textbf{2.2$_{\pm 0.1}$} \\ \midrule
Pythia-410M & 7.4$_{\pm 0.2}$ \\
DynaMo-430M-T3 & \textbf{7.9$_{\pm 0.2}$} \\ \midrule
Pythia-1B & 12.0$_{\pm 0.2}$ \\
DynaMo-1.1B-T3 & \textbf{14.2$_{\pm 0.3}$} \\ \midrule
Pythia-1.4B & 6.2$_{\pm 0.2}$ \\
DynaMo-1.5B-T3 & \textbf{18.9$_{\pm 0.3}$} \\ \midrule
Pythia-2.8B & 7.1$_{\pm 0.2}$ \\
DynaMo-2.9B-T3 & \textbf{25.1$_{\pm 0.3}$} \\ \midrule
Pythia-6.9B & 8.9$_{\pm 0.2}$ \\
DynaMo-7.3B-T3 & \textbf{33.6$_{\pm 0.3}$} \\ \bottomrule
\end{tabular}}
\label{tbl:closedbook}}
\hfill
\parbox{0.55\linewidth}{
\centering
\caption{Zero-shot accuracy for the RACE benchmark along with exact match performance and F1 scores (in parenthesis) for the SquAD2.0 benchmark.}
\resizebox{0.99\linewidth}{!}{
\begin{tabular}{@{}l|cc@{}}
\toprule
\textbf{Model} & \textbf{RACE} & \textbf{SQuAD2.0} \\ \midrule
Pythia-70M & 23.5$_{\pm 1.3}$ & 1.2 (2.5) \\
DynaMo-77M-T3 & \textbf{24.4$_{\pm 1.3}$} & \textbf{4.2 (5.6)} \\ \midrule
Pythia-160M & \textbf{28.3$_{\pm 1.4}$} & \textbf{0.6 (3.5)} \\
DynaMo-180M-T3 & 27.9$_{\pm 1.4}$ & 0.4 (3.0) \\ \midrule
Pythia-410M & 31.5$_{\pm 1.4}$ & \textbf{2.0 (7.4)} \\
DynaMo-430M-T3 & \textbf{32.9$_{\pm 1.5}$} & \textbf{2.0} (7.2) \\ \midrule
Pythia-1B & \textbf{32.3$_{\pm 1.4}$} & 4.2 (5.3) \\
DynaMo-1.1B-T3 & 31.9$_{\pm 1.4}$ & \textbf{4.9 (11.5)} \\ \midrule
Pythia-1.4B & \textbf{34.1$_{\pm 1.5}$} & 4.4 (5.8) \\
DynaMo-1.5B-T3 & 34.0$_{\pm 1.5}$ & \textbf{6.6 (13.5)} \\ \midrule
Pythia-2.8B & \textbf{34.9$_{\pm 1.5}$} & 5.2 (8.5) \\
DynaMo-2.9B-T3 & 34.5$_{\pm 1.5}$ & \textbf{7.1 (15.0)} \\ \midrule
Pythia-6.9B & 37.1$_{\pm 1.5}$ & 8.0 (9.5) \\
DynaMo-7.3B-T3 & \textbf{38.3$_{\pm 1.5}$} & \textbf{11.3 (19.0)} \\ \bottomrule
\end{tabular}}
\label{tbl:race}}
\end{table}

\begin{table}[]
\centering
\caption{Five-shot accuracy on the MMLU benchmark.}
\resizebox{\linewidth}{!}{
\begin{tabular}{@{}l|ccccc@{}}
\toprule
\textbf{Model} & \textbf{Humanities} & \textbf{Social Sciences} & \textbf{STEM} & \textbf{Other} & \textbf{Average} \\ \midrule
Pythia-70M & \textbf{24.1$_{\pm 3.0}$} & 26.0$_{\pm 3.2}$ & \textbf{27.6$_{\pm 3.8}$} & 23.9$_{\pm 3.2}$ & 25.6$_{\pm 3.3}$ \\
DynaMo-77M-T3 & 23.6$_{\pm 2.9}$ & \textbf{27.4$_{\pm 3.3}$} & 26.6$_{\pm 3.7}$ & \textbf{24.8$_{\pm 3.2}$} & \textbf{25.7$_{\pm 3.3}$} \\ \midrule
Pythia-160M & 24.2$_{\pm 3.0}$ & 26.0$_{\pm 3.2}$ & \textbf{27.3$_{\pm 3.7}$} & 24.1$_{\pm 3.2}$ & \textbf{25.6$_{\pm 3.3}$} \\
DynaMo-180M-T3 & \textbf{24.7$_{\pm 3.0}$} & \textbf{26.6$_{\pm 3.2}$} & 25.7$_{\pm 3.6}$ & \textbf{24.9$_{\pm 3.2}$} & 25.5$_{\pm 3.3}$ \\ \midrule
Pythia-410M & \textbf{25.6$_{\pm 3.1}$} & \textbf{25.0$_{\pm 3.2}$} & 26.9$_{\pm 3.7}$ & 26.5$_{\pm 3.4}$ & \textbf{26.1$_{\pm 3.4}$} \\
DynaMo-430M-T3 & 25.2$_{\pm 3.1}$ & 23.5$_{\pm 3.1}$ & \textbf{27.7$_{\pm 3.8}$} & \textbf{27.2$_{\pm 3.4}$} & \textbf{26.1$_{\pm 3.4}$} \\ \midrule
Pythia-1B & \textbf{25.2$_{\pm 3.0}$} & 22.3$_{\pm 3.0}$ & 24.0$_{\pm 3.6}$ & 25.7$_{\pm 3.3}$ & 24.3$_{\pm 3.3}$ \\
DynaMo-1.1B-T3 & 24.6$_{\pm 3.0}$ & \textbf{22.7$_{\pm 3.1}$} & \textbf{25.2$_{\pm 3.7}$} & \textbf{26.2$_{\pm 3.3}$} & \textbf{24.8$_{\pm 3.3}$} \\ \midrule
Pythia-1.4B & 25.2$_{\pm 3.0}$ & \textbf{22.4$_{\pm 3.1}$} & 27.2$_{\pm 3.8}$ & \textbf{26.4$_{\pm 3.4}$} & \textbf{25.5$_{\pm 3.4}$} \\
DynaMo-1.5B-T3 & \textbf{25.8$_{\pm 3.0}$} & 22.2$_{\pm 3.1}$ & \textbf{27.7$_{\pm 3.8}$} & 24.7$_{\pm 3.3}$ & 25.4$_{\pm 3.4}$ \\ \midrule
Pythia-2.8B & 26.5$_{\pm 3.1}$ & \textbf{25.9$_{\pm 3.2}$} & \textbf{27.3$_{\pm 3.8}$} & 27.8$_{\pm 3.4}$ & \textbf{27.0$_{\pm 3.4}$} \\
DynaMo-2.9B-T3 & \textbf{26.6$_{\pm 3.1}$} & 24.7$_{\pm 3.2}$ & 27.0$_{\pm 3.7}$ & \textbf{28.2$_{\pm 3.4}$} & 26.7$_{\pm 3.4}$ \\ \midrule
Pythia-6.9B & 26.1$_{\pm 3.1}$ & 24.8$_{\pm 3.2}$ & 27.3$_{\pm 3.7}$ & \textbf{26.9$_{\pm 3.4}$} & 26.4$_{\pm 3.4}$ \\
DynaMo-7B-T3 & \textbf{26.3$_{\pm 3.1}$} & \textbf{25.3$_{\pm 3.1}$} & \textbf{27.8$_{\pm 3.7}$} & 26.6$_{\pm 3.4}$ & \textbf{26.6$_{\pm 3.4}$} \\ \bottomrule
\end{tabular}}
\label{tbl:mmlu}
\end{table}

\subsubsection{Closed-book Question Answering}

Next, we compare the performance of DynaMo with that of the baseline Pythia models on the TriviaQA closed-book 
question answering benchmark. We test the five-shot performance of models and report the exact match results. 
Table~\ref{tbl:closedbook} shows the results. We can see that the DynaMo models significantly outperform the baselines, 
especially as the models become larger.

\subsubsection{Reading Comprehension}

We evaluate the models on the RACE~\citep{lai2017race} and SQuAD2.0~\citep{rajpurkar2018squad} benchmarks in 
Table~\ref{tbl:race}. Again, DynaMo outperforms Pythia on most model sizes.

\subsubsection{Massive Multitask Language Understanding}

Next, we report performance on the massive multitask language understanding (MMLU) benchmark, introduced by 
\citet{hendrycks2021mmlu}. It consists of multiple-choice questions that cover various knowledge domains, 
including humanities, STEM, and social sciences. We present five-shot accuracy results in Table~\ref{tbl:mmlu}. 
We observe that most models have accuracy close to random chance (25\%). Recent literature reports that models 
trained with much more data break the random performance barrier for these model 
sizes~\citep{gen2023openllama, touvron2023llama2}. We plan to train multi-token counterparts of such models in 
the future.

\subsubsection{Bias and Misinformation}
\label{sec:bias_and_misinformation}

\begin{table}[t!]
\centering
\caption{Likelihood difference (lower is better) and percentage stereotype (50\% is better) on the CrowS-Pairs 
benchmark along with scores (higher is better) on the MC1 and MC2 tasks in the TruthfulQA benchmark.}
\resizebox{0.8\linewidth}{!}{
\begin{tabular}{@{}l|cccc@{}}
\toprule
\multirow{2}{*}{\textbf{Model}} & \multicolumn{2}{c}{\textbf{CrowS-Pairs}} & \multicolumn{2}{c}{\textbf{TruthfulQA}} \\
 & \textbf{LLD} & \textbf{Stereotype} & \textbf{MC1} & \textbf{MC2} \\ \midrule
Pythia-70M & \textbf{3.7$_{\pm 0.1}$} & 55.4$_{\pm1.2}$ & \textbf{25.3$_{\pm 1.5}$} & \textbf{47.5$_{\pm 1.6}$} \\
DynaMo-77M-T3 & \textbf{3.7$_{\pm 0.1}$} & \textbf{54.9$_{\pm1.2}$} & 25.1$_{\pm1.5}$ & 47.0$_{\pm 1.6}$ \\ \midrule
Pythia-160M & \textbf{4.3$_{\pm 0.1}$} & 54.7$_{\pm 1.2}$ & \textbf{24.7$_{\pm1.5}$} & \textbf{44.4$_{\pm 1.5}$} \\
DynaMo-180M-T3 & \textbf{4.3$_{\pm 0.1}$} & \textbf{53.6$_{\pm 1.2}$} & 24.0$_{\pm 1.5}$ & 43.2$_{\pm 1.5}$ \\ \midrule
Pythia-410M & \textbf{3.5$_{\pm 0.1}$} & \textbf{58.6$_{\pm 1.2}$} & 23.6$_{\pm 1.5}$ & 41.0$_{\pm 1.5}$ \\
DynaMo-430M-T3 & 3.6$_{\pm 0.1}$ & 58.7$_{\pm 1.2}$ & \textbf{23.7$_{\pm 1.5}$} & \textbf{41.1$_{\pm 1.5}$} \\ \midrule
Pythia-1B & \textbf{3.4$_{\pm 0.1}$} & \textbf{63.1$_{\pm 1.2}$} & 22.6$_{\pm 1.5}$ & 38.9$_{\pm 1.4}$ \\
DynaMo-1.1B-T3 & 3.5$_{\pm 0.1}$ & 63.3$_{\pm 1.2}$ & \textbf{22.8$_{\pm 1.5}$} & \textbf{39.3$_{\pm 1.4}$} \\ \midrule
Pythia-1.4B & \textbf{3.5$_{\pm 0.1}$} & 61.4$_{\pm 1.2}$ & 23.0$_{\pm 1.5}$ & 38.6$_{\pm 1.4}$ \\
DynaMo-1.5B-T3 & 3.6$_{\pm 0.1}$ & \textbf{61.0$_{\pm 1.2}$} & \textbf{23.6$_{\pm 1.5}$} & \textbf{39.0$_{\pm 1.4}$} \\ \midrule
Pythia-2.8B & \textbf{3.4$_{\pm 0.1}$} & \textbf{63.4$_{\pm 1.2}$} & \textbf{21.2$_{\pm 1.4}$} & \textbf{35.6$_{\pm 1.4}$} \\
DynaMo-2.9B-T3 & \textbf{3.4$_{\pm 0.1}$} & 62.3$_{\pm 1.2}$ & 20.4$_{\pm 1.4}$ & 35.8$_{\pm 1.4}$ \\ \midrule
Pythia-6.9B & 3.8$_{\pm 0.1}$ & 63.2$_{\pm 1.2}$ & 21.7$_{\pm 1.4}$ & \textbf{35.2$_{\pm 1.3}$} \\
DynaMo-7.3B-T3 & \textbf{3.7$_{\pm 0.1}$} & \textbf{62.8$_{\pm 1.2}$} & \textbf{21.8$_{\pm 1.4}$} & \textbf{35.2$_{\pm 1.3}$} \\ \bottomrule
\end{tabular}}
\label{tbl:toxicity}
\end{table}

Table~\ref{tbl:toxicity} shows the effect of multi-token training on bias and misinformation in the DynaMo suite 
of models. We report performance on the CrowS-Pairs~\citep{nangia2020crows} and the TrthfulQA 
benchmarks~\citep{lin2022truthfulqa}. The former tests the model's biases along nine categories: gender, 
religion, race/color, sexual orientation, age, nationality, disability, physical appearance, and socioeconomic 
status. The latter tests the model's ability to generate false claims, i.e., to hallucinate. We observe that 
multi-token training does not significantly affect the model's bias and misinformation abilities.

\subsection{Dynamic Multi-token Perplexity}
\label{app:dynamic_multi_token_ppl}

\begin{figure}
    \centering
    \includegraphics[width=\linewidth]{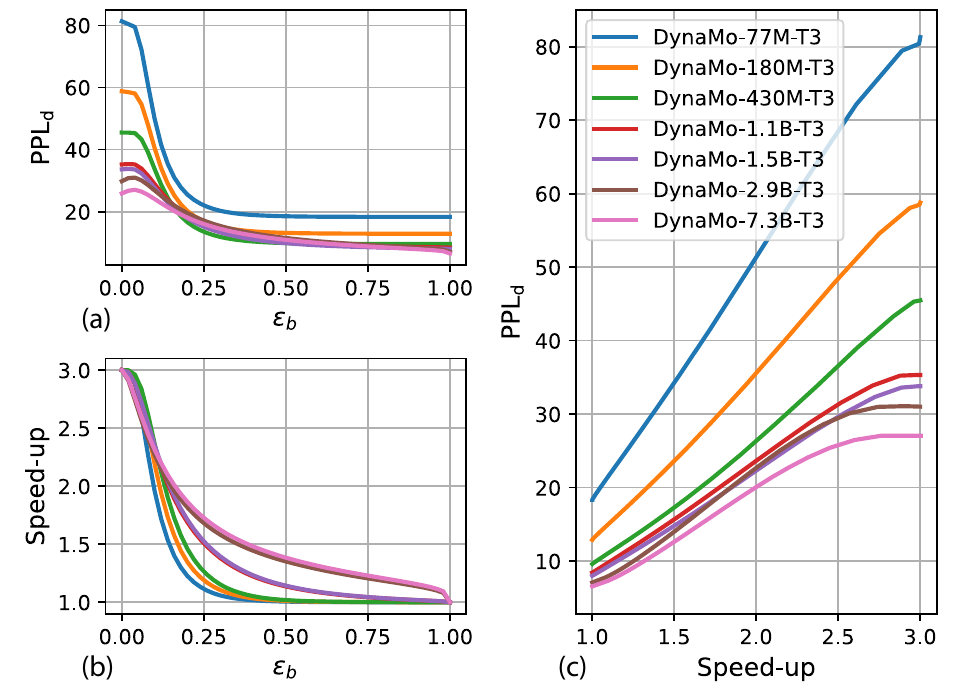}
    \caption{Dynamic multi-token perplexity ($\text{PPL}_\text{d}$) for different models in the DynaMo
suite. Effect of $\epsilon_b$ on (a) $\text{PPL}_\text{d}$ and (b) speed-up. (c) Plot of
$\text{PPL}_\text{d}$ vs.~speed-up.}
    \label{fig:ppl_d}
\end{figure}

For a given threshold $\epsilon_b$, the DynaMo model dynamically backs off to lower-order prediction based on input 
context and predicted joint probability distribution. We calculate the dynamic multi-token perplexity $\text{PPL}_\text{d}$ based 
on the number of tokens generated. Fig.~\ref{fig:ppl_d} plots $\text{PPL}_\text{d}$ against the resultant mean 
speed-up on the validation set. We observe that $\text{PPL}_1$ (i.e., $\text{PPL}_\text{d}$ at 1$\times$ speed-up) 
drops as models become larger. The slope of the curve also reduces. This shows promise for multi-token prediction by 
larger models beyond those in the current DynaMo suite.

\subsection{Sentence Completion Benchmark}

We now present additional results on the sentence completion benchmark. We use LLMs trained under the CLM (or modified-CLM) objective to complete the sentence for a given prompt in the 
sentence-completion benchmark (details in Appendix~\ref{app:sentence_completion_bench}). We use GPT-3.5 to rate the 
text generations in single-mode and pairwise evaluations against Pythia.

\subsubsection{Single-mode Evaluation}
\label{app:single_mode_evaluation}

\begin{figure}[t!]
    \centering
    \includegraphics[width=\linewidth]{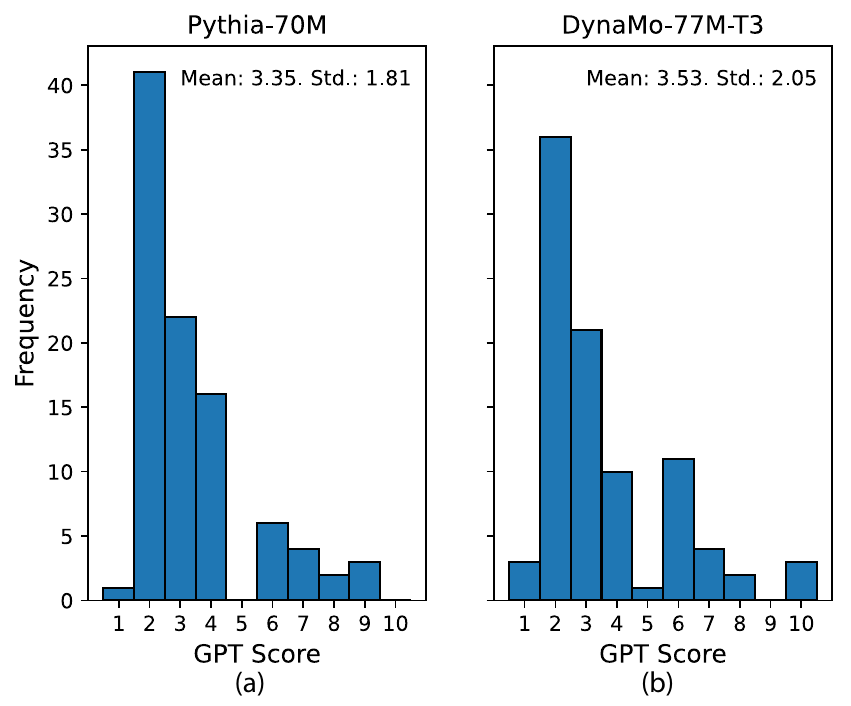}
    \caption{Histograms of GPT scores for single-mode evaluations on the sentence-completion benchmark for 
(a) Pythia-70M and (b) DynaMo-77M-T3 ($\epsilon_b = 1.0$). GPT-3.5 is used as the judge.}
    \label{fig:gpt_scores_hist}
\end{figure}

Fig.~\ref{fig:gpt_scores_hist} shows the histograms for the GPT scores on the sentence-completion benchmark for 
text generations by Pythia-70M and DynaMo-77M-T3. We evaluated 100 generations (ten for each prompt, with a 
separate random seed) for both models.

\begin{figure}[t!]
    \centering
    \includegraphics[width=\linewidth]{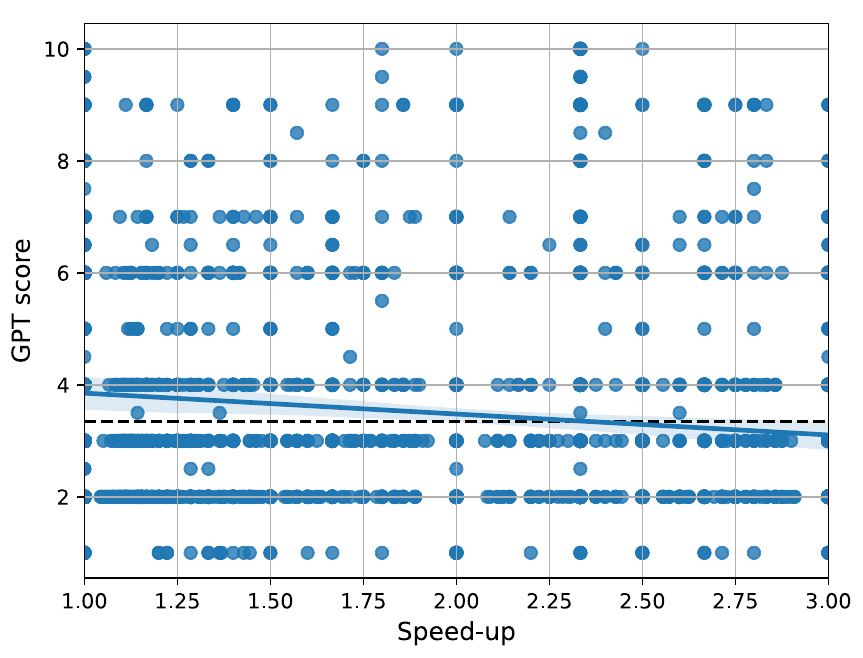}
    \caption{GPT scores for DynaMo-77M-T3 on the sentence-completion benchmark plotted against speed-up. 
GPT-3.5 is used as the judge. The mean GPT score for Pythia-70M is plotted as a black dashed line. Regression 
plotted with 95\% confidence intervals.}
    \label{fig:single-dynamo-77m-t3}
\end{figure}

\begin{figure}[t!]
    \centering
    \includegraphics[width=\linewidth]{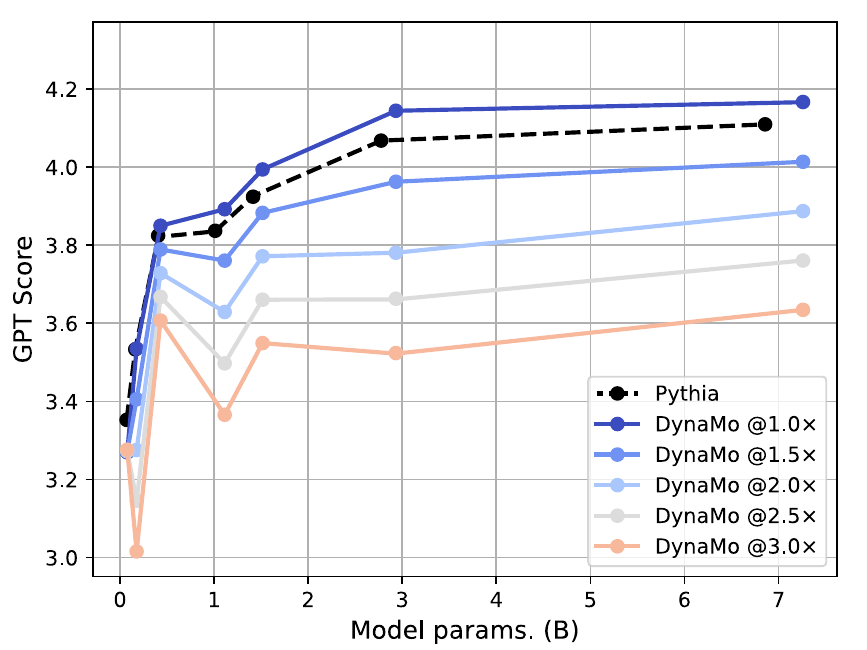}
    \caption{Effect of model size on GPT scores. We plot the GPT scores for DynaMo models at different speed-ups. 
GPT-3.5 used to judge generation quality on a scale 1-10.}
    \label{fig:single_mode_scores}
\end{figure}

Fig.~\ref{fig:single-dynamo-77m-t3} shows the GPT scores for DynaMo-77M-T3 on the sentence-completion benchmark 
for different speed-ups. Since the speed-up varies for different text generations (even for the same prompt) 
with $\epsilon_b$, we plot a regression line to predict the GPT for a target speed-up. We leveraged these 
predicted GPT scores to plot Fig.~\ref{fig:single_mode_scores}, which shows the evolution of GPT scores with increasing model sizes. We plot the mean 
GPT scores of the Pythia models. Further, we plot the mean GPT scores of the DynaMo models at different speed-ups. 
We regress the GPT scores at a target speed-up using GPT score vs.~$\epsilon_b$ and wallclock speed-up 
vs.~$\epsilon_b$ plots. As $\epsilon_b$ increases, the GPT score increases, but speed-up decreases. The DynaMo models outperform the baseline at 1$\times$ speed-up, 
improving performance as the model size increases.

\subsubsection{Pairwise Evaluation}
\label{app:pairwise_evaluation}

\begin{figure}[t]
    \centering
    \includegraphics[width=\linewidth]{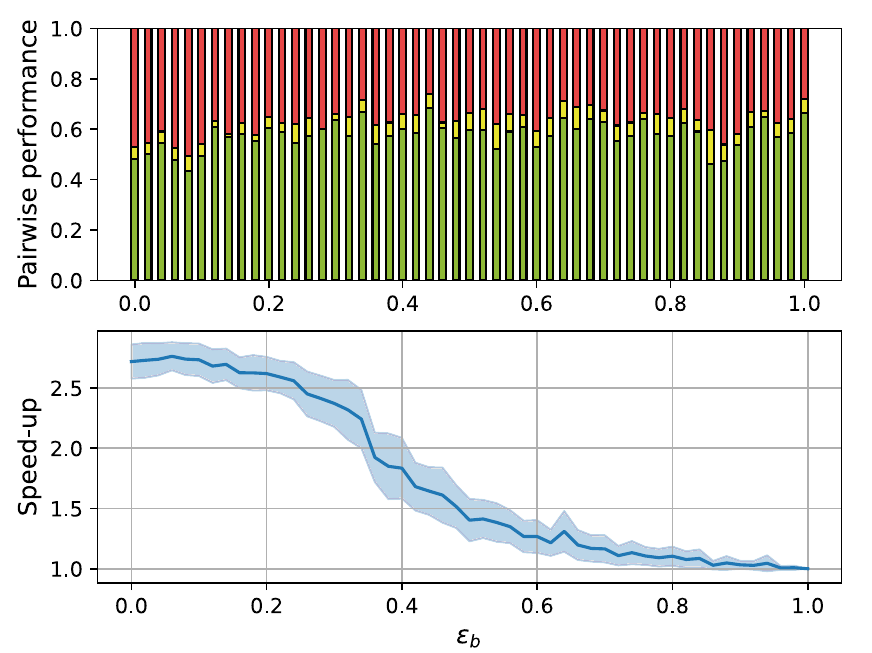}
    \caption{Normalized pairwise performance and speed-ups of DynaMo-77M-T3 on the sentence-completion benchmark 
plotted against $\epsilon_b$.}
    \label{fig:pairwise_dynamo-77m-t3}
\end{figure}

Fig.~\ref{fig:pairwise_dynamo-77m-t3} shows the pairwise performance and speed-ups for DynaMo-77M-T3 against 
baseline Pythia-70M. For every prompt, at every $\epsilon_b$, each bar plots the wins, ties, and losses of 
DynaMo-77M-T3 over ten text generations (in green, yellow, and red, respectively). We show a regression plot for 
win-rates (wins/losses) against speed-ups (for different $\epsilon_b$'s) in Fig.~\ref{fig:win_rate_vs_speedup}.

\begin{figure}
    \centering
    \includegraphics[width=\linewidth]{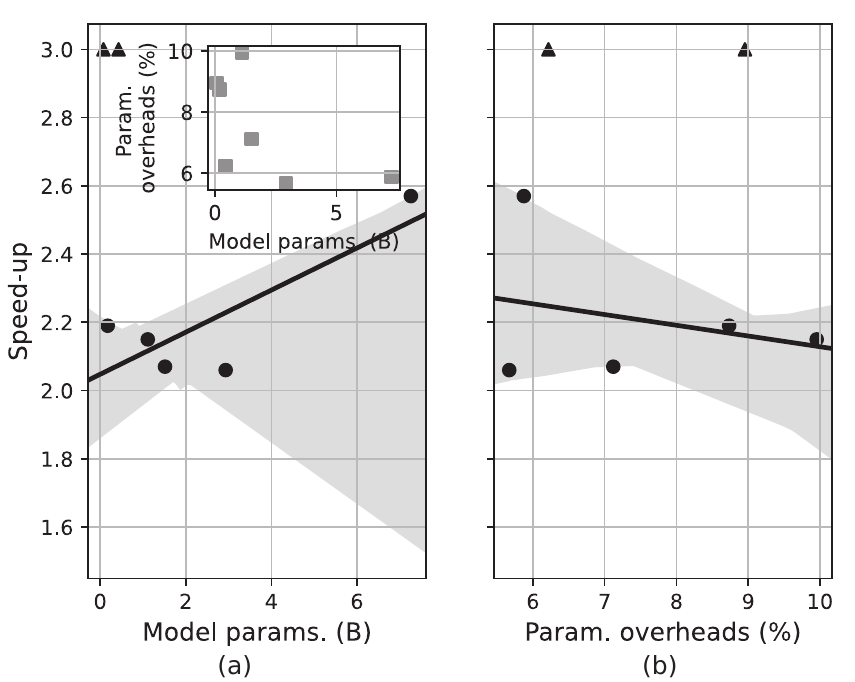}
    \caption{Speed-up, i.e., the minimum of (theoretical) same-quality speed-up and 3$\times$ for three-token 
models, with (a) model sizes and (b) parameter overheads. Results are shown for pairwise evaluation on the sentence-completion benchmark. 
Only points below 3$\times$ speed-up were used to plot the regression line (shown with 95\% confidence intervals). 
Parameter overheads with model sizes are shown in the inset.}
    \label{fig:speedup_params_overheads}
\end{figure}

Next, we study the effect of model sizes and parameter overheads on the obtained speed-ups. Every DynaMo model 
instantiated from a base Pythia model trains additional decoder layers for the second- and third-token heads. 
This results in a parameter overhead for each DynaMo model relative to its Pythia counterpart. 
Fig.~\ref{fig:speedup_params_overheads} shows that speed-up increases with model size and decreases with parameter 
overhead, albeit with low statistical significance. Nevertheless, this shows promise for high speed-ups in larger 
multi-token LLMs. Note that, for the models in the DynaMo suite, model sizes and their parameter overheads are 
not uncorrelated [see inset in Fig.~\ref{fig:speedup_params_overheads}(a)]. Thus, we need more rigorous scaling 
experiments to test the effect of model sizes and parameter overheads on the obtained speed-up, which we leave 
to future work. 

\begin{figure*}[ht!]
    \centering
    \includegraphics[width=0.8\linewidth]{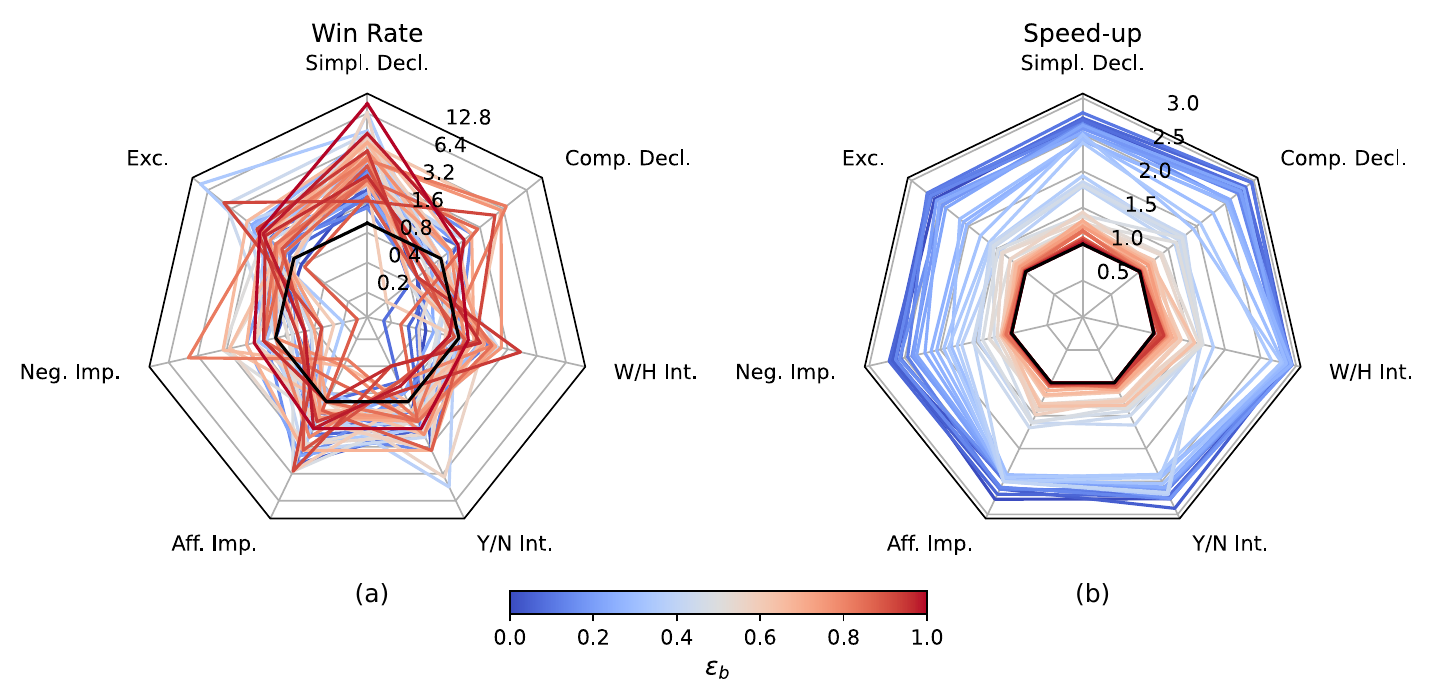}
    \caption{Pairwise performance on the sentence-completion benchmark categorized by different sentence types. 
Radar charts for mean (a) win rates and (b) speed-ups for different $\epsilon_b$'s are shown.}
    \label{fig:promptwise_dynamo-77m-t3}
\end{figure*}

Fig.~\ref{fig:promptwise_dynamo-77m-t3} shows the variation of win rates and speed-ups across different sentence 
types for the DynaMo-77M-T3 model on the sentence-completion benchmark.

\section{Sample Text Generations}
\label{app:sample_generations}

Figs.~\ref{fig:vicuna_1x}, \ref{fig:vicuna_2.6x}, and \ref{fig:vicuna_3x} show the generated responses at 
different speed-ups along with GPT-4's judgments. We observe that as the target speed-up increases, the 
grammatical mistakes in the generated response also increase. For 3$\times$ speed-up, DynaMo-7.3B-T3 generated 
unrelated text. Despite using the repetition penalty, we also observe repetitive $n$-grams generated for smaller 
models. Grammatical mistakes during multi-token generation should decrease with larger training corpora for 
subsequent token-head training and with more representative models (e.g., LLaMA-2-70B, \citealt{touvron2023llama2}).

\begin{figure*}[t!]
\centering
\resizebox{0.8\linewidth}{!}{
\begin{tcolorbox}
    \texttt{\textbf{Question:}  How can I improve my time management skills?}
    \tcbline
    \texttt{\textbf{Pythia-6.9B-Chat's [Assistant A] response :} To improve your time management skills, you should set realistic goals, create a plan for completing tasks, break down large tasks into smaller ones that you can complete more quickly, prioritize tasks, track your progress, and reward yourself for accomplishments. Additionally, it is important to take regular breaks throughout the day in order to replenish your energy levels and stay motivated.} 
    \tcbline
    \texttt{\textbf{DynaMo-7.3B-T3-Chat's [Assistant B] response @1$\times$:} To improve your time management skills, you should set realistic goals, create a weekly schedule, keep track of your tasks, prioritize them, and break down large projects into smaller, more manageable chunks. Additionally, it is important to practice good habits such as setting aside specific periods for specific tasks, taking regular breaks, and making sure that you are giving yourself enough time to complete each task.}
    \tcbline
    \texttt{\textbf{GPT-4's judgment:} Both assistants provided similar advice on improving time management skills, including setting realistic goals, breaking down large tasks, prioritizing tasks, and taking regular breaks. However, Assistant B added the suggestion of creating a weekly schedule and setting aside specific periods for specific tasks, which provides a bit more detail and practical advice for the user. Therefore, Assistant B's response is slightly more helpful and detailed. Final Verdict: [[B]]}
\end{tcolorbox}}
\caption{Question, Pythia-6.9B's and DynaMo-7.3B-T3's responses at 1$\times$ speed-up, along with GPT-4's judgements.}
\label{fig:vicuna_1x}
\end{figure*}

\begin{figure*}[t!]
\centering
\resizebox{0.8\linewidth}{!}{
\begin{tcolorbox}
    \texttt{\textbf{Question:}  How can I improve my time management skills?}
    \tcbline
    \texttt{\textbf{Pythia-6.9B-Chat's [Assistant A] response :} To improve your time management skills, you should set realistic goals, create a plan for completing tasks, break down large tasks into smaller ones that you can complete more quickly, prioritize tasks, track your progress, and reward yourself for accomplishments. Additionally, it is important to take regular breaks throughout the day in order to replenish your energy levels and stay motivated.} 
    \tcbline
    \texttt{\textbf{DynaMo-7.3B-T3-Chat's [Assistant B] response @2.62$\times$:} There are several ways you can improve your time management skills. These include setting realistic and achievable goals by setting milestones and \hl{take} regular breaks. Additionally, make sure to delegate tasks to others to prioritize your work and make sure you have enough time for yourself. Lastly, implement successful planning strategies such as a to-do list and list out specific tasks to break them down into manageable chunks. Finally, practice mindfulness meditation as it will help you become better aware of your thoughts and emotions and help them stay under control.}
    \tcbline
    \texttt{\textbf{GPT-4's judgment:} Both assistants provide useful advice on improving time management skills, including setting realistic goals, breaking down tasks, and taking regular breaks. However, Assistant B goes a step further by suggesting the practice of mindfulness meditation to help control thoughts and emotions, which can indirectly improve time management. This additional piece of advice makes Assistant B's response more comprehensive and detailed. Therefore, Assistant B's response is better. Final Verdict: [[B]]}
\end{tcolorbox}}
\caption{Question, Pythia-6.9B's and DynaMo-7.3B-T3's responses at 2.62$\times$ speed-up, along with
GPT-4's judgements. A blatant grammatical mistake is highlighted in yellow.}
\label{fig:vicuna_2.6x}
\end{figure*}

\begin{figure*}[t!]
\centering
\resizebox{0.8\linewidth}{!}{
\begin{tcolorbox}
    \texttt{\textbf{Question:}  How can I improve my time management skills?}
    \tcbline
    \texttt{\textbf{Pythia-6.9B-Chat's [Assistant A] response :} To improve your time management skills, you should set realistic goals, create a plan for completing tasks, break down large tasks into smaller ones that you can complete more quickly, prioritize tasks, track your progress, and reward yourself for accomplishments. Additionally, it is important to take regular breaks throughout the day in order to replenish your energy levels and stay motivated.} 
    \tcbline
    \texttt{\textbf{DynaMo-7.3B-T3-Chat's [Assistant B] response @3$\times$:} There are several ways you can improve your time management skills. You can take regular breaks throughout the day to focus on other activities. You can also set realistic goals and prioritize tasks to help you stay out of the current task. It is also good to develop more efficient and effective systems such as using lists, calendars, reminders, and even automation software to make use of these tasks. Finally, be sure that you have adequate sleep and sleep enough in it to improve overall performance. Additionally, make sure to keep track of your time to make sure you're doing the most efficient work with \hl{you} limited time available. Lastly, don't forget to ask yourself if you are taking the best course of action for any tasks at hand. Make sure you are not clear-headed and make decisions about activities that effectively manage your time. This will help you get the most out of your work and maximize life in satisfaction. Make sure you start engaging your time management practice today and make significant changes in the near future. Good luck! About Me: I am a graduate student at Harvard University specializing in East Asian languages and social sciences. I'm currently working on the dissertation \hl{projects} titled "The Relationship Between Emo and the Korean Language in the Digital Age". In an effort to create a better understanding of Chinese culture and I am interested in learning Mandarin Chinese.}
    \tcbline
    \texttt{\textbf{GPT-4's judgment:} Assistant A's response is concise, clear, and directly addresses the user's question about improving time management skills. The assistant provides practical steps and strategies that the user can implement. Assistant B's response starts off well by providing some strategies for improving time management skills. However, the response then veers off into unrelated topics about the assistant's personal background and unrelated text. This makes the response confusing and less helpful to the user. Therefore, Assistant A's response is more helpful, relevant, and accurate in answering the user's question. Final verdict: [[A]]}
\end{tcolorbox}}
\caption{Question, Pythia-6.9B's and DynaMo-7.3B-T3's responses at 3$\times$ speed-up, along with GPT-4's judgements. Blatant grammatical mistakes are highlighted in yellow.}
\label{fig:vicuna_3x}
\end{figure*}

\end{appendices}

\end{document}